\documentclass{article} % For LaTeX2e
\usepackage{iclr2027_conference,times}

\usepackage{amsmath,amsfonts,bm}

\def\eqref#1{equation~\ref{#1}}
\def\1{\bm{1}}

\DeclareMathAlphabet{\mathsfit}{\encodingdefault}{\sfdefault}{m}{sl}
\SetMathAlphabet{\mathsfit}{bold}{\encodingdefault}{\sfdefault}{bx}{n}

\definecolor{citecolor}{HTML}{2980b9}
\definecolor{linkcolor}{HTML}{c0392b}
\usepackage[pagebackref=true,breaklinks=true,colorlinks,bookmarks=false,citecolor=citecolor,linkcolor=linkcolor]{hyperref}

\usepackage{url}
\usepackage{booktabs}
\usepackage{graphicx}
\usepackage{dsfont}

\usepackage{amsmath}
\usepackage{amssymb}
\usepackage{algorithm}
\usepackage{algorithmic}
\usepackage{makecell}
\usepackage{dsfont}
\usepackage{wrapfig}
\usepackage{enumitem}

\title{A World Model for Assembly Sequence Planning}

\newif\ifarxiv
\arxivtrue

\author{Xilun Chen \\ The Chinese University of Hong Kong, Shenzhen
\And
Roozbeh Mottaghi \\ University of Washington}

\definecolor{revisecolor}{HTML}{000000}
\newcommand{\revise}[1]{{\color{revisecolor}#1}}

\ifarxiv
  \iclrpreprintcopy   % authors + head "Preprint. Under review." (defined in the .sty)
\fi
\begin{document}

\maketitle

\begin{abstract}
This paper proposes a world model for planning assembly sequences for LEGO structures. The model predicts placement feasibility by looking ahead in a learned latent space, without invoking a physical simulator inside the look-ahead search. The primary technical challenge lies in judging whether committing to a placement leaves the remaining structure buildable, which requires reasoning over future steps. To address this, we represent the assembly with a multi-granularity heterogeneous graph, in which a master node aggregates the overall geometry of the assembled structure as a set of anchor tokens, while a brick-level sub-graph carries the interlocking topology and the static attributes of each brick. A latent dynamics model advances the anchor tokens locally after each placement, which lets the world model roll the geometry forward over several steps and anticipate whether a candidate placement obstructs the assembly of the remaining bricks. We validate the framework on held-out test structures from the StableText2Lego~\citep{brickgpt} benchmark, comparing it against ablations and recent assembly-sequence methods. Our planner achieves near-oracle solutions while issuing the fewest exact-oracle calls per brick and avoiding the later-infeasible placements incurred by the baselines. 
\ifarxiv
Project page: \url{https://assembly-world-model.github.io/}
\fi
\end{abstract}

\section{Introduction}
\label{sec:intro}
\vspace{-0.3cm}
An essential task for embodied AI and robotics is automated assembly: enabling one or more robots to compose a set of individual parts into a target structure. The key to this problem is \emph{assembly sequence planning} (ASP), determining the order in which parts are placed, as this order decides whether the assembly is physically realizable~\citep{asap}. We study ASP on brick assembly, a reproducible testbed with readily available components and existing benchmarks; brick assembly is easily scaled to hundreds of parts and exposes the core physical constraints of other assembly problems, such as interlocking connectivity, per-step stability, and collision-free reachability~\citep{paca, brickbybrick, brickgpt}.

A core challenge in ASP is the need to \emph{look ahead} over future steps~\citep{s2a}, as a placement that looks feasible now may leave the remaining structure unbuildable later. Existing methods either ignore look-ahead and place parts greedily one step at a time~\citep{paca}, or obtain look-ahead by querying physical ground truth through time-consuming, unscalable simulation and motion planning~\citep{s2a, grace}. This leaves a gap in designing a feasibility judge that can look ahead over many steps without paying the per-step cost of a physical oracle.

\begin{figure}[t]
\centering
\includegraphics[width=\columnwidth]{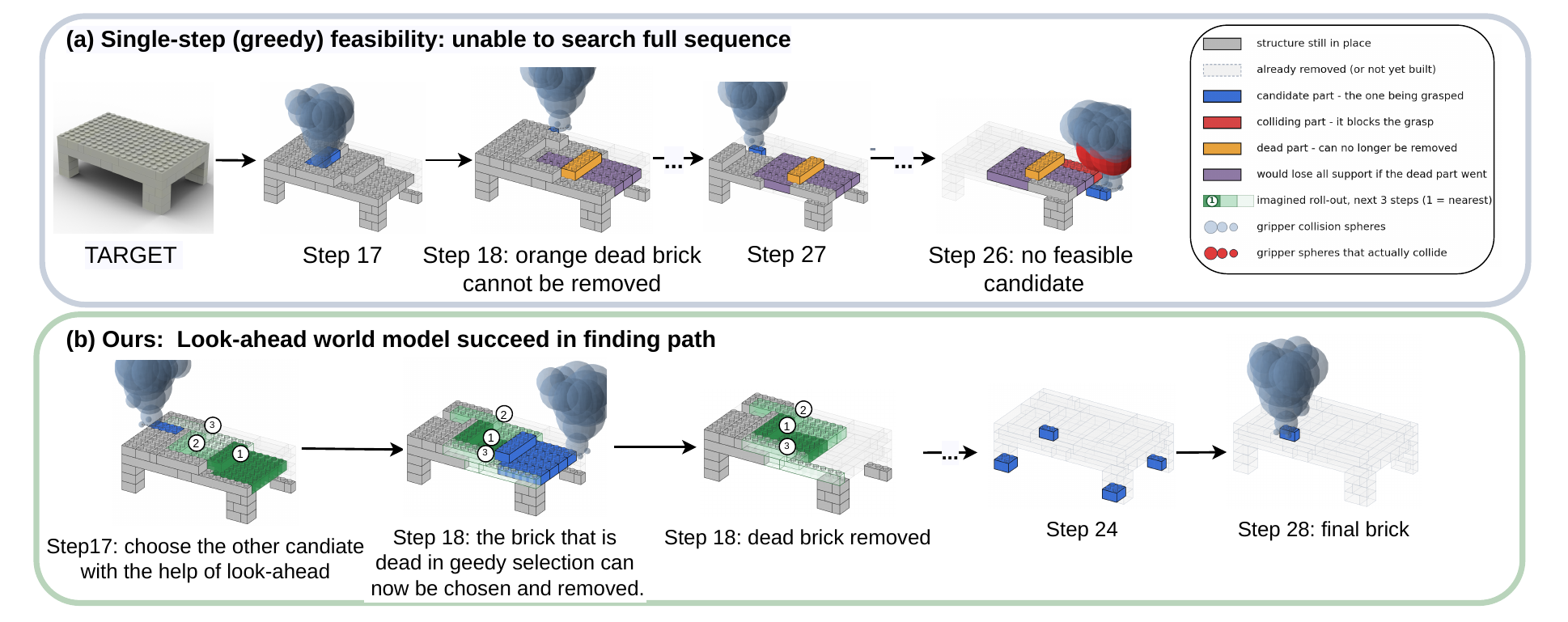}
\vspace{-0.7cm}
\caption{A LEGO structure with a dead-end assembly order(upper) vs a correct one(lower). Our world model succeed by rolling feasibility forward and avoid committing an early candidate disassembly(blue candidate in upper row) which would blocks a later one(orange dead brick). The spheres in the figure denote the gripper's collision geometry.}
\vspace{-0.3cm}
\label{fig:teaser}
\end{figure}

We address this gap with a \emph{manipulation-level world model} and solve the brick assembly sequence problem by assembly-by-disassembly, recursively removing bricks from the completed structure to obtain a feasible order (Figure~\ref{fig:teaser}). We train the world model to predict whether a placement is feasible and to imagine how that feasibility evolves as further bricks are removed, so that the planner can look many steps ahead before committing to a brick. To keep this prediction faithful, the model represents the evolving assembly as a multi-granularity heterogeneous graph, reasoning jointly about the global geometry that governs collisions and the local connectivity that governs stability. Concretely, two learned predicates read the overall geometry and inner brick structure and two deterministic gates handle the structure-independent reachability and press checks, whereas an encoder summarizes the built geometry into a set of latents and a dynamics model advances the latents after each rollout. The planner rolls this world model forward to score candidates and commit a removal order, which we detail in Section~\ref{sec:method}.

On the StableText2Lego benchmark, this look-ahead raises buildable-assembly success to $90.5\%$ on the full held-out test split, above greedy ordering ($78.9\%$) and above a single-step exact-oracle predicate ($85.7\%$), while reducing oracle calls by more than half. Under a shared exact static-stability oracle, it outperforms recent assembly-sequence methods (WorkBenchMark~\citep{workbenchmark}, ASAP~\citep{asap}, H-WM~\citep{hwm}, PhysWorld~\citep{physworld} and OR solver~\citep{culbertson}) by a wide margin. These results show that a learned world model can provide multi-step feasibility look-ahead at oracle-level quality without the per-step cost that makes exact look-ahead intractable. 

In summary, we contribute: 1) a manipulation-level world model to imagine and evaluate assembly steps; 2) a multi-granularity heterogeneous graph structured model serving a collision judge and a stability judge that operate at different structural levels; 3) a brick assembly sequence planner that outperforms prior assembly-sequence methods and ablations under a shared physical ruler.

\vspace{-0.3cm}
\section{Related Work}
\vspace{-0.3cm}
\label{sec:related}

% ------------------------------------------------------------
\noindent\textbf{Assembly Sequence Planning.}
Assembly sequence planning determines the placement order that makes a structure
physically realizable, and prior work spans a spectrum from greedy ordering to
look-ahead that relies on an exact-physics oracle. At one end, methods solve for an order while encoding physics only as a deterministic mask or a single-step validity check: placing bricks under a data-free action mask~\citep{paca}, filtering an indefinite action space with a learned validity predictor~\citep{brickbybrick, assemblycomplete}, imitating ground-truth orders with graph transformers~\citep{icraasp, treesba}, masking illegal actions~\citep{somacube}, or solving an optimal sequential decision with dynamic programming or reinforcement learning~\citep{orasp}. Robotic-building methods learn placement policies~\citep{learningtobuild} or reason over sequence and contact jointly~\citep{multilevel, euphoria}, yet treat gripper reachability as a downstream check or future work. Others predict single-step feasibility directly, such as directional grasp feasibility~\citep{grn} or collision and clearance surrogates~\citep{fastron, clearancenet}, but these are static and reversible and never roll feasibility forward over the irreversible multi-step process we address. Moving toward look-ahead, another line evaluates each candidate against physical ground truth at deployment through simulation and motion planning~\citep{s2a, grace, asap}, search-based disassembly planners further accelerate this exact evaluation through intermediate-state reuse and directional blocking graphs~\citep{statebased}. This line of method is exact yet pays a per-step oracle cost that scales poorly. When focusing on brick assembly tasks specifically, prior systems are largely confined to generation~\citep{brickgpt, stable} or to deterministic reconstruction planners that check reachability by explicit collision testing~\citep{frombricks, workbenchmark, apexmr}, none performing learned multi-step feasibility look-ahead. We address this gap by learning the physical feasibility judgment and rolling it forward in latent space, so the planner anticipates multi-step consequences without paying a per-step oracle or assembly re-construction cost at inference.

% ------------------------------------------------------------
\noindent\textbf{World Models.}
Learning a model to replace an expensive physical simulator goes back to learned interaction engines and graph-network simulators~\citep{interactionnet, gns, worldmodels}, which latent world models extend to multi-step imagination for planning~\citep{planet, dreamerv3, tdmpc2, iris}. Their look-ahead, however, targets a task reward or visual future rather than physical feasibility, and generic latent rollouts do not by themselves preserve physical validity~\citep{jepawm, hamiltonian}. Object-centric and graph-structured world models make look-ahead compositional~\citep{cswm, fiocwm, objectzero}, but the graph-world-model survey's physical role collects only generic collision, friction, and particle models, none addressing assembly legality~\citep{gwmsurvey}. Recent methods that use a world model as a critic over candidate actions did not consider physical feasibility as well~\citep{gpc, wav}; we share this critic mechanism but predict physical placement feasibility rather than a task reward or a rendered future, which is precisely the physical validity that generic latent rollouts fail to preserve. 

% ------------------------------------------------------------
\noindent\textbf{Graph Representation and Stability Prediction.}
Our world model couples a geometry encoder over a multi-granularity heterogeneous
graph with a learned stability predictor. The anchor-token encoder follows the
point-proxy design of PoinTr~\citep{pointr}, and its multi-level structure builds on hierarchical graph pooling and heterogeneous graph networks~\citep{diffpool, hgt}. The stability predictor is a learned structural judge, an ability with ample precedent, from predicting stability from appearance~\citep{visualstability} to regressing structural response with graph networks~\citep{gnnstress, gnndisp}. We adopt these representations for a different role: rather than encoding a shape for reconstruction or scoring stability in isolation, our multi-granularity graph lets a single world model reason jointly about global collision geometry (through the master-node anchor latents) and local interlocking connectivity (through the brick sub-graph), and exposes stability as a readout that the planner queries at every look-ahead step.

\section{Assembly Planning with a World Model}
\label{sec:method}
\vspace{-0.3cm}
% ------------------------------------------------------------
\subsection{Problem Formulation}
\label{sec:problem}
\vspace{-0.3cm}
We study assembly sequence planning for brick assembly: given a target structure, we must order its bricks so that every placement is physically executable by an embodiment and leaves a stable structure. We solve this with a learned world model that predicts placement feasibility and looks ahead over multiple steps. We first formalize what makes a step feasible. The input is a target structure specified as a set of $M$ axis-aligned bricks $B = \{b_1, \dots, b_M\}$ on an integer stud grid, where each brick is given by its footprint dimensions and its discrete position. The output is an ordered assembly sequence: a list of steps, each of which places a single brick or a rigidly-coupled group of bricks (of size $K \le 5$) onto the partially-built structure, using a specified manipulation motion and a designated gripped brick. The placements are executed by a dual-arm embodiment (VAMP-MR~\citep{vampmr}). At any step we write $b_{\text{built}}$ for the aggregate of already-placed bricks and $b_{\text{active}}$ for the brick or group added in that step.

A step is admissible only if it can be physically executed and leaves a stable structure. We describe a step by an action $a = (b_{\text{active}}, m, b_{\text{grip}})$, where $b_{\text{active}}$ is the brick or rigidly-coupled group being moved, $m$ is the manipulation motion (top-down placement or bottom-up insertion), and $b_{\text{grip}}$ is the brick within the group that is in direct contact with the gripper. The action is applied to the current aggregate $b_{\text{built}}$. We formalize feasibility as a conjunction of five predicates: an action $a$ is feasible if and only if
\begin{equation}
\begin{aligned}
\mathrm{feasible}(a) \iff\;
& \underbrace{\mathrm{press}(a) \,\wedge\, \mathrm{reach}(a)}_{\text{deterministic gates}} \,\wedge\, \underbrace{\mathrm{collision}(a)}_{\text{learned collision}} \\[2pt]
&\wedge\; \underbrace{\mathrm{stable}(b_{\text{built}}) \,\wedge\, \mathrm{stable}(b_{\text{built}} \setminus b_{\text{active}})}_{\text{learned stability}} .
\end{aligned}
\label{eq:feasible}
\end{equation}
Each action expands into a finite set of discrete gripper configurations that realize it, differing in the arm assignment, the approach side, and the tool offset. Manipulation-dependent predicates quantify existentially over this set: $\mathrm{press}(a)$ holds if there exists a configuration whose gripper force point, the stud through which force is applied, lies within the admissible bounds; $\mathrm{reach}(a)$ holds when some configuration admits inverse-kinematics solutions for its motion waypoints; and $\mathrm{collision}(a)$ holds when some configuration admits a collision-free placement against the assembled structure. The stability predicate $\mathrm{stable}(\cdot)$ requires the aggregate to remain stable under gravity, checked both before and after $b_{\text{active}}$ is detached so that the structure is stable throughout. The first two predicates are structure-independent, depending only on the action and its target pose, whereas the latter three depend on the assembled geometry. Our objective is to find a disassembly sequence in which every removal step is feasible, so that the reversed order is a feasible assembly sequence. 

% ------------------------------------------------------------
\vspace{-0.3cm}
\subsection{Structural Abstraction}
\label{sec:repr}
\vspace{-0.3cm}
A brick structure admits a natural graph abstraction in which each node is a brick and each edge encodes a stud-level interlock between two vertically adjacent bricks. We build our system on this abstraction while separating the \emph{planner} from the \emph{world model}, each operating on its own view of the graph. The planner reasons over a single flat brick connection graph and solves the assembly order by \emph{disassembly}: starting from the complete target and repeatedly removing one brick or rigidly-coupled group at a time until nothing remains, after which the reversed removal order gives a valid assembly sequence. Planning in this direction is what makes the problem tractable, since it constrains the search to orders that reach the known target and lets each step reason about detaching from a fully specified structure rather than searching forward over exponentially many partial builds~\citep{s2a}. The world model, in contrast, maintains a hierarchical heterogeneous graph, in which a master node summarizes the assembled structure while a brick-level sub-graph is attached beneath it.

\begin{figure}[t]
\centering
\includegraphics[width=\columnwidth]{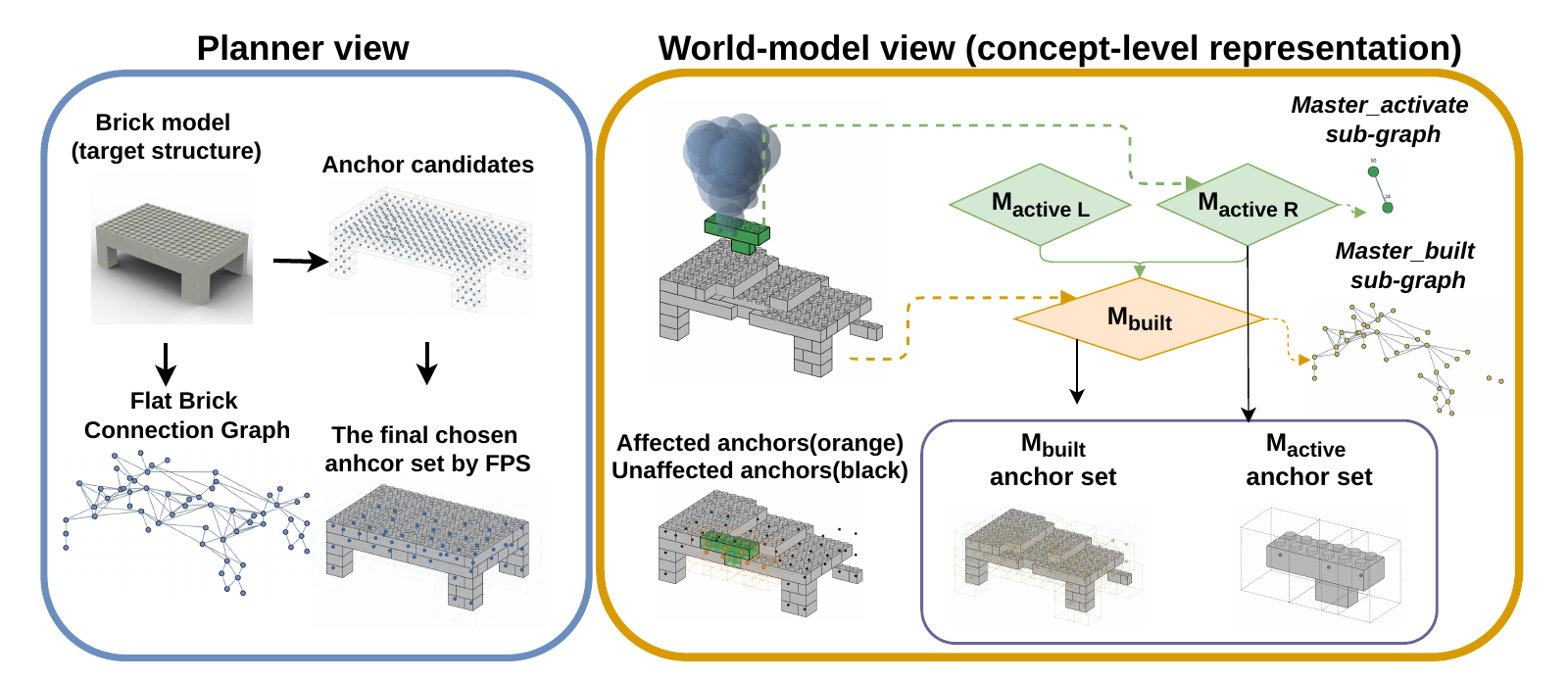}
\vspace{-0.7cm}
\caption{\textbf{Structural abstraction:} flat brick connection graph and anchor set selection (planner view) vs. hierarchical heterogeneous latent graph and anchor tokens renewing (world-model view).}
\vspace{-0.7cm}
\label{fig:abstraction}
\end{figure}

Since the stability predicates depend only on the interlocking topology among the placed bricks, whereas collision-freeness depends only on the exposed surface geometry of the assembled aggregate, the heads that predict feasibility in Section~\ref{sec:problem} consume information at different structural levels. We therefore give the world model a hierarchical heterogeneous graph: a master node $M_{\text{built}}$ aggregates the overall geometry of the assembled aggregate $b_{\text{built}}$ into a geometric latent, while a brick-level sub-graph beneath it carries the interlocking topology together with a per-brick attribute vector  (Figure~\ref{fig:abstraction}). Latent geometry avoids tracking the full pose of every brick throughout a rollout. The attribute vector is an $11$-dimensional summary of each brick's footprint, mass, and current support relations within the structure (its height, and the number of bricks it rests on and supports); the same vector is shared by the encoder, the dynamics model, and the stability predictor, and we give its full specification in supplementary material.

 As shown in Figure~\ref{fig:abstraction}, the geometric latent carried by the master node is not a single vector but a set of \emph{anchor tokens}, attached to a fixed spatial location within the structure. The anchors are generated by farthest-point sampling over all occupied cells of the target structure, and the anchor set remains fixed throughout the assembly. Each anchor carries a token summarizing the geometry of its local neighborhood. We describe each cell in that neighborhood by a \emph{cell token}: whether it is occupied, which of its six faces are exposed, and the attribute vector of the brick occupying it. These anchor tokens are produced by the encoder and rolled forward as the geometric state of the world model.
 
% ------------------------------------------------------------
\vspace{-0.3cm}
\subsection{World-Model Architecture}
\label{sec:heads}
\begin{figure}[t]
\centering
\includegraphics[width=\columnwidth]{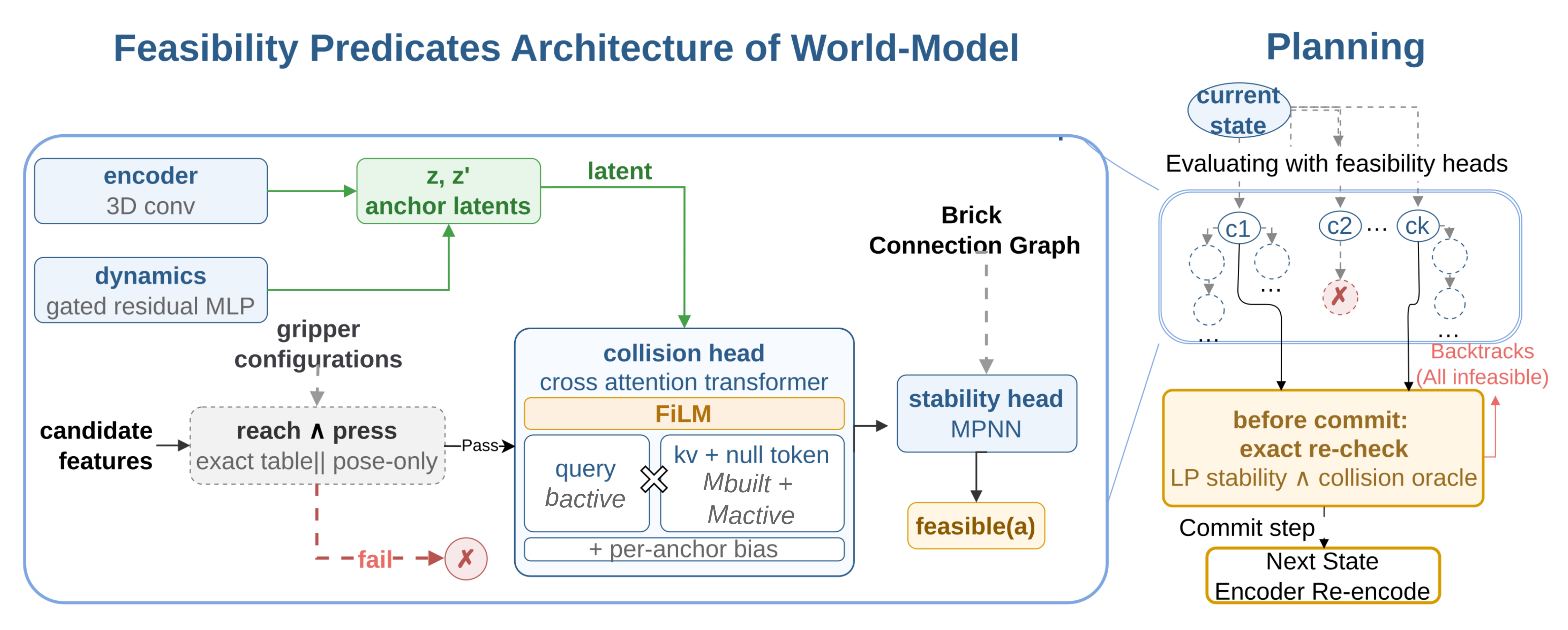}
\vspace{-0.7cm}
\caption{\textbf{Feasibility predictor} of world model (left) and planner (right). Left: a candidate passes through the feasibility predicates; blue blocks are learned modules, the dashed block is the deterministic reach,$\wedge$,press gate, and the green arrow marks the anchor latents, read only by the collision head. Right: one step of the disassembly search, with solid arrows for committed steps and dashed arrows for the imagined depth-$d$ rollout; a red cross marks a first-layer dead end, and the best-scoring candidate is exactly re-checked before commit. \emph{LP} denotes the exact linear-program stability check applied at commit time.} 
\label{fig:pipeline}
\vspace{-0.5cm}
\end{figure}

\vspace{-0.3cm}
The world model realizes the five predicates of Section~\ref{sec:problem} (Figure~\ref{fig:pipeline}, left). Two of them, $\mathrm{press}(a)$ and $\mathrm{reach}(a)$, are deterministic gates over structure-independent geometry, while $\mathrm{collision}(a)$ and the two stability terms are learned. We describe the encoder and the dynamics model first, then the two gates, and finally the two learned predicates $\mathrm{collision}$ and $\mathrm{stable}$. 

\paragraph{Encoder.}
The encoder maps each anchor's neighborhood to a latent representation. After each committed removal, the imagined state is replaced by the actual structure and the latents are recomputed. During look-ahead rollouts between commits, the dynamics model below instead updates the latents incrementally without re-encoding the structure. For anchor $i$, it computes $z_i = \mathrm{Enc}(\mathbf{c}_i)$ from the cell tokens $\mathbf{c}_i$ in the cube centered at $i$ (Section~\ref{sec:repr}), by embedding each cell token, applying local 3D convolutions over the cube, and pooling them with a mask, so that $z_i$ summarizes only the geometry in its own neighborhood.
The encoder is trained self-supervised. The training signal comes from an MLP decoder discarded when inferring. From each cell token the decoder $\mathrm{Dec}$ must recover the cell's neighborhood occupancy $o_i$:
$\mathcal{L}_{\mathrm{enc}} = \mathrm{BCE}\big(\mathrm{Dec}(z_i),\, o_i\big)$.
The encoder is then frozen when training other heads.

\paragraph{Dynamics.}
During a rollout we do not re-encode the whole structure after every placement. Instead, a dynamics model renews only the latents of the anchors near the placement. Writing $z_i$ for the latent of anchor $i$ and $z_i'$ for its value after the update, it applies a gated residual update
\begin{equation}
\begin{aligned}
z_i' = z_i + \mathds{1}\!\left[i \in \mathcal{A}(a)\right] \cdot
\mathrm{MLP}\big(\left[\,z_i,\; \psi_i(a)\,\right]\big),
\end{aligned}
\end{equation}
where $\mathcal{A}(a)$ is the set of anchors whose neighborhood is geometrically
affected by action $a$, $\mathds{1}[\cdot]$ is the indicator that gates the update to
those anchors, $[\,\cdot\,,\,\cdot\,]$ denotes concatenation, and $\psi_i(a)$ is a fixed featurization of the placed brick relative to anchor $i$: the brick's occupancy
footprint within that anchor's cube concatenated with its attribute vector. The final
MLP layer is zero-initialized, and is trained self-supervised on $\mathcal{A}(a)$: each updated latent $z_i'$ must decode to the true next occupancy $o_i^{+}$ and stay close to the true next-state latent $\bar z_i$ through a relative term with  $\lambda_z{=}10$ and $\mathrm{sg}$ stop-gradient applied to $\bar z_i$.
\begin{equation}
\mathcal{L}_{\mathrm{dyn}} = \mathrm{BCE}\big(\mathrm{Dec}(z_i'), \,o_i^{+}\big)
+ \lambda_z \frac{\lVert z_i' - \mathrm{sg}(\bar z_i)\rVert^2}{\lVert \mathrm{sg}(\bar z_i)\rVert^2},\quad i\in\mathcal{A}(a),
\end{equation}

With the dynamics head, the model does \emph{not} read the occupancy grid, the exposed-face set, or the per-brick contact counts of the next state, but infers the latent change from the removed brick alone.

\paragraph{Deterministic gates.}
The two gates are training-free exact checks that depend only on the action and its target pose, not on the assembled structure, so they are evaluated first and filter out candidates cheaply. Because the motion waypoints are given in closed form over a finite pose domain, we precompute $\mathrm{reach}(a)$, the existence of inverse-kinematics solutions for those waypoints, into an exact lookup table answered in a single query. $\mathrm{press}(a)$ analytically checks that the press point, the stud through which the gripper applies the assembly force, lies within its admissible bounds.

\paragraph{Collision.}
The $\mathrm{collision}$ predicate predicts whether placing the candidate group
$b_{\text{active}}$ onto $b_{\text{built}}$ can be carried out without geometric
collision. As an action expands into a finite set of discrete gripper configurations
$\mathcal{G}(a)$ (Section~\ref{sec:problem}), the head scores the whole set in a single
forward pass: for each configuration $g \in \mathcal{G}(a)$ it outputs a feasibility
logit $\ell_g$ and, as an auxiliary output, a signed distance $d_g$ between
the moved group and the structure.

We instantiate $\mathrm{collision}$ as an edge biased cross attention module. Collision depends on how the placed group and the arm sweeping it into position interact with the exposed geometry of the assembled structure. We therefore treat each candidate gripper configuration as a query and let it attend to the anchor latents of $M_{\text{built}}$ and $M_{\text{active}}$, which serve as keys and values. This allows each configuration to focus on the small subset of anchors most likely to contend for the same space. We augment the keys and values with a learnable null token, ensuring that attention remains well defined when $M_{\text{built}}$ contains few or no anchors. Because the relative position of an anchor with respect to the placed group provides a strong geometric prior for contact, we inject this information explicitly as a per anchor bias on the attention logits rather than requiring the attention module to recover it from the latent representations alone. Importantly, this bias is computed once per anchor, rather than separately for every configuration and anchor pair. Finally, the two motion modes induce qualitatively different collision geometries, namely top-down placement and bottom-up insertion, so we condition the module on the motion $m$ using FiLM ~\ref{sec:problem}. The collision module is trained on top of the frozen encoder. 
The model is trained supervisedly using an exact collision oracle ~\ref{sec:planning} to generate ground truth signed distance $\hat d_g$ and logit $ y_g$ with $\lambda_d{=}0.3$:
\begin{equation}
\mathcal{L}_{\mathrm{col}} = \mathrm{BCE}(\ell_g,\, y_g)
  \;+\; \lambda_d\,\mathrm{Huber}_{\delta=1}(d_g,\, \hat d_g).
\end{equation}

\paragraph{Stability.}
The stability predicate evaluates $\mathrm{stable}(b_{\text{built}})$ and $\mathrm{stable}(b_{\text{built}} \setminus b_{\text{active}})$, requiring the sub-assembly to stand both before and after the group is detached. Unlike $\mathrm{collision}$, the stability head does not read the anchor latents; it is a message-passing neural network (MPNN) that operates directly on the brick adjacency graph, predicting a per-brick instability logit $s_b$, which is passed through a sigmoid to output an instability probability, and we declare a state stable when every brick's predicted instability probability stays below a calibrated threshold $\tau$. We build the training dataset from the StableLego brick assemblies with per-brick stability labels~\citep{stablelego}. Since the structures in the data are complete, their unstable bricks are mostly unsupported, whereas our disassembly planner more often faces overloaded studs and sparse remnants that tip over. We therefore add self-generated partial structures by repeatedly stripping bricks from complete brick assemblies while keeping all remaining bricks in structures connected to other bricks or the ground. We then form a supplementary set of unstable remnants and label them by an exact stability solver. The stability head is trained jointly on the combined dataset with labeled ground-truth stability logit $y_b$:
$\mathcal{L}_{\mathrm{stab}} = \mathrm{BCE}_{w}(s_b,\, y_b).$

% ------------------------------------------------------------
\vspace{-0.3cm}
\subsection{Planning with the World Model}
\label{sec:planning}
\vspace{-0.3cm}

We plan in the disassembly direction (Figure~\ref{fig:pipeline}, right): starting from the complete target structure, at each step we enumerate the candidates that can currently be removed (a single brick or a rigidly-coupled group), filter them with the feasibility predicates of Section~\ref{sec:heads}, and commit one removal before repeating until the structure is empty; executing the committed removals in reverse yields the assembly sequence. Among the candidates that pass the predicates, the planner commits the highest-scoring $b_{\text{active}}$ that is not a dead end, where the score is produced by the look-ahead below and ties are broken by a deterministic candidate order. If a state has no committable candidate, the planner backtracks to the previous commit and tries its next-best candidate.

 For each passing candidate, the world model performs a depth-$d$ look-ahead in the latent space, a layered rollout advanced by the dynamics model. The roots of the rollout are the currently feasible candidates of the current state; at each subsequent layer, every imagined placement is expanded only along the bricks geometrically adjacent to the group just removed, since removing a group can only alter the feasibility of its immediate neighbors. All placements within a layer are scored with a single forward pass of the collision head, and placements that reach the same state through different orders are merged, which keeps the number of candidates evaluated per step in the single digits. The score of a root is the total number of feasible continuations exposed across its whole subtree: the more feasible continuations a candidate opens up, the higher its score. A candidate is a dead end if its successor state is non-empty yet exposes no feasible continuation at all. A dead end at the first layer vetoes the root directly, whereas at deeper layers an infeasible branch simply contributes nothing to the score rather than ruling the root out.

Before a candidate is committed, the selected action is re-checked by two exact solvers. The stability check is a linear-program relaxation of the StableLego force-equilibrium analysis~\citep{stablelego}, which is provably sound and matches the original verdicts under a per-instance certificate (more details in supplementary material). The collision oracle verifies the placement through analytic inverse-kinematics and geometric collision computation; to speed it up, the collision head's per-placement scores from the look-ahead are reused so that the oracle first examines the placements the head ranks most feasible. A candidate that fails either check is recorded as a false pass and skipped, so every step of the returned assembly sequence is exactly verified.

\vspace{-0.3cm}
\section{Experiments}
\label{sec:exp}
\vspace{-0.3cm}
The goal of our experiments is to verify whether 1) look-ahead brings a measurable benefit over greedy single-step planning; 2) our world-model based planner with look-ahead is strong both in ablations and in comparison with state-of-the-art baselines; and 3) our collision head needs to read the full geometry latent anchor set.

% =====================================================================
% Setup
% =====================================================================
\vspace{-0.3cm}
\subsection{Datasets and Setup}
\label{sec:exp-setup}
\vspace{-0.3cm}
\noindent\textbf{Data and representation.} We evaluate on the StableText2Lego corpus~\citep{brickgpt}, a collection of stud-level LEGO structures spanning 21 object categories. Every structure is built from a library of \emph{eight brick sizes} ($1{\times}1$, $1{\times}2$, $1{\times}4$, $1{\times}6$, $1{\times}8$, $2{\times}2$, $2{\times}4$, $2{\times}6$); all bricks are axis-aligned rectangular cuboids. Each object may appear as several distinct \emph{structures}, where the same shape comes with alternative brick layouts. We apply disassembly sequence planning on these structures.

We split the corpus at the \emph{object} level: $24{,}711$ objects for training and $697$ for validation. For evaluation, we use the official held-out test split ($2{,}851$ objects, $4{,}785$ structures) \revise{for the ablations, and draw a fixed, stratified sample of $1{,}200$ structures from it to compare with external baselines,} preserving the split's category and structure-size distribution. 

\noindent\textbf{Embodiment.} All planning runs use a single, fixed embodiment: the dual-arm LEGO workcell distributed with VAMP-MR~\citep{vampmr} (its \texttt{dual\_gp4} environment), i.e.\ two 6-DoF Yaskawa Motoman GP4 arms, each carrying that repository's LEGO end-of-arm tool (a custom LEGO gripper with a wrist-mounted RGB-D camera). Robot geometry is the sphere-based collision model shipped with the same environment.

\noindent\textbf{Exact references.} Two exact, deterministic components provide ground truth. (i) An \emph{exact collision oracle}: closed-form analytic inverse kinematics plus sphere-based geometric collision checking over the manipulation waypoint chain. (ii) \emph{Exact statics}: the StableLego force-equilibrium program~\citep{stablelego}, which we solve with \textbf{HiGHS}~\citep{highs} as a linear program, whose tightness we verify per instance with a complementarity certificate (experiment result in supplementary material). The collision oracle serves as the label source for the learned collision head, while exact statics provides the labels for training the stability head (Section~\ref{sec:heads}). Both components serve as the commit-time re-check inside planning (Section~\ref{sec:method}).

% =====================================================================
% Metrics
% =====================================================================
\vspace{-0.3cm}
\subsection{Metrics} 
\vspace{-0.2cm}
\label{sec:exp-metrics}

Metrics of full method performance:
\begin{itemize}[topsep=0.2pt, nosep]
\item\textbf{Succ.} (success rate): Fraction of evaluable structures whose recorded sequence is feasible at every step and removes all bricks: each step must satisfy $\mathrm{feasible}(a)$ (Eq.~\ref{eq:feasible}) instantiated with the \emph{exact} collision oracle and \emph{exact} statics; any infeasible step is a failure. The criterion is identical across methods: $k$ is the number of structures found a feasible disassembly sequence and $n$ is the number of total evaluated structures.
  \item\textbf{Orc/br.} (oracle calls per brick): The median number of per-structure exact collision-oracle calls a method spends per brick. This is a measure of how much exact checking a method needs.
  \item\textbf{Imagined/br.} (imagined states per brick): The median number of per-structure latent look-ahead states the method advances per brick. This is a measure of search effort.
  \item\textbf{Wall-clock.} The median per-structure wall-clock time(s).
  \item\textbf{McNemar.} As methods time out on different structures, success is compared on paired subsets, using the exact-binomial McNemar test. We report the true $p$-value or $p<10^{-k}$.
\end{itemize}

Metric of collision head analysis:
\begin{itemize}[topsep=1pt,nosep]
  \item\textbf{FP@R99.} False-positive rate at $99\%$ recall: at the threshold that still accepts $99\%$ of the feasible cells, the fraction of infeasible cells it lets through; lower is better.
\end{itemize}
% =====================================================================
% Q0 · Gallery
% =====================================================================

\begin{figure}[t]
\centering
\includegraphics[width=\columnwidth]{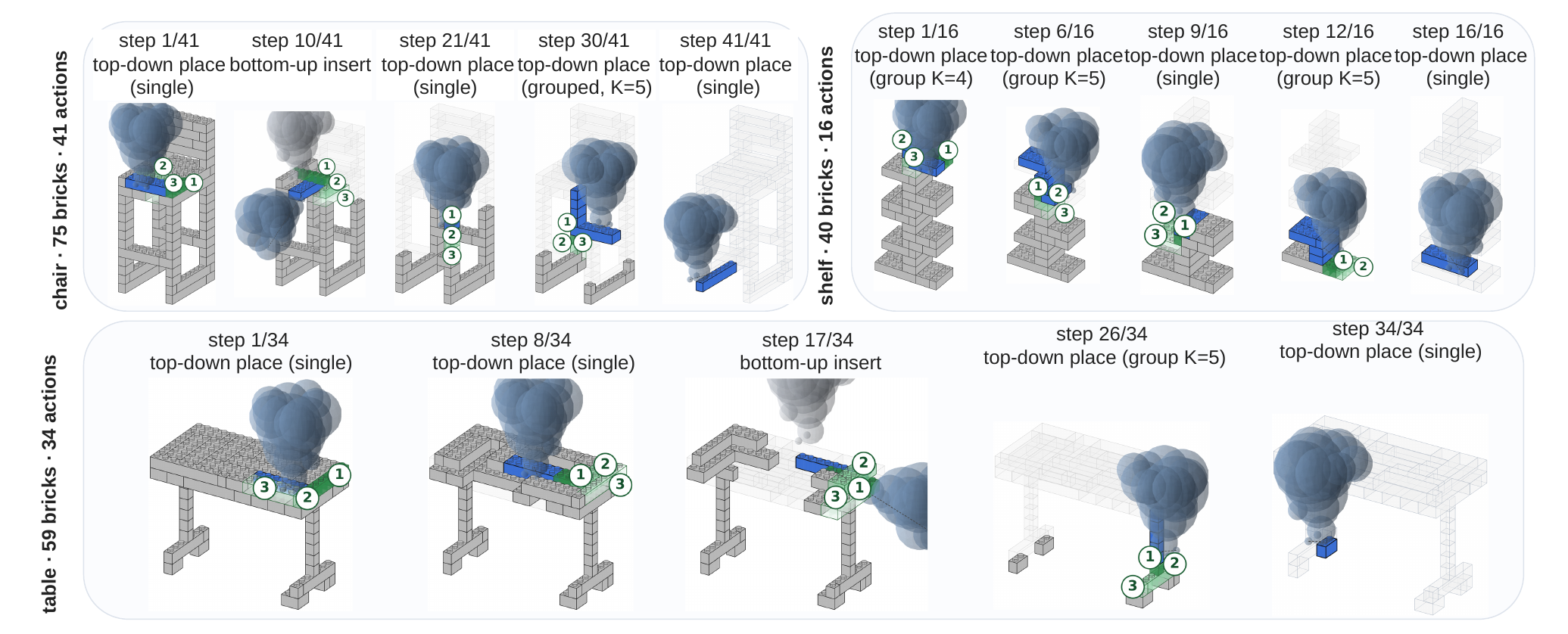}
\vspace{-0.7cm}
\caption{\textbf{Qualitative Results.} Qualitative disassembly galleries for the three models. In each keyframe the current candidate brick is blue and the look-ahead bricks are green, with darker shades marking deeper look-ahead steps. For bottom-up insertion the holding arm is shown in steel-blue and the top-pressing arm that stabilizes the sub-assembly in light-gray. } 
\vspace{-0.5cm}
\label{fig:gallery}
\end{figure}

% =====================================================================
% Q1 · Look-ahead depth
% =====================================================================
\subsection{Look-Ahead Evaluation}
\label{sec:exp-lookahead}
\vspace{-0.3cm}
We isolate the effect of look-ahead by fixing the predicate and the planner and varying only the roll-out depth $d$ (Table~\ref{tab:lookahead}), which shows that a greedy single-step planner is not sufficient. Success improves with depth \revise{and look-ahead halves the exact-oracle calls per brick.} Its cost is computation and wall-clock time, which grow with the number of imagined states expanded per brick. \revise{$d{=}5$ yields no significant gain over $d{=}3^\ast$ ($p{=}0.22$) at roughly eight times as many imagined states. The result indicates that d=1 cannot see the traps, which need at least two steps to expose. The gain plateaus at d=3, and we adopt it for our full-method cross-comparison.}

\begin{table}[t]
\centering
\caption{\textbf{Look-ahead depth ablation analysis} \revise{on the held-out test split.} McNemar compares each row with the row above. \revise{Rows $0$--$3$ are paired on the $n{=}4177$ structures on which no depth times out ($0\!\to\!3$: $p{<}10^{-20}$); $3^\ast$ and $5$ on the $n{=}988$ structures of at most $60$ bricks on which $d{=}5$ does not time out.}}
\label{tab:lookahead}
\small
\begin{tabular}{lcccccc}
\toprule
Depth & $n$ & Succ. & McNemar & Orc/br. & Imagined/br. & Wall-clock (s) \\
\midrule
0        & \revise{4177} & \revise{0.903} & ---                 & \revise{1.21} & 0   & \revise{45}  \\
1        & \revise{4177} & \revise{0.921} & \revise{$p{<}10^{-5}$}       & 0.48 & 7   & \revise{29}  \\
2        & \revise{4177} & \revise{0.936} & \revise{$p{<}10^{-7}$}       & \revise{0.48} & \revise{32}  & \revise{76}  \\
3 (ours) & \revise{4177} & \revise{0.938} & \revise{$p{=}0.37$}          & \revise{0.51} & \revise{126} & \revise{242} \\
$3^\ast$ & \revise{988}  & \revise{0.979} & ---                 & \revise{0.58} & \revise{85}  & \revise{35}  \\
5        & \revise{988}  & \revise{0.983} & \revise{$p{=}0.22$}          & \revise{0.65} & \revise{665} & \revise{177} \\
\bottomrule
\end{tabular}
\vspace{-0.6cm}
\end{table}

% =====================================================================
% Q2 · The world model replaces the exact solver (MAIN)
% =====================================================================
\vspace{-0.3cm}
\subsection{Cross-Method Comparison}
\label{sec:exp-main}
\vspace{-0.3cm}
We study whether the world model can act as the feasibility predicate during planning without losing solution quality. Under an identical commit-time verifier and look-ahead depth, we build a ladder of methods, each adding new components over the previous one, named as \textsc{predicate used}/\textsc{whether using rollout} \revise{(Table~\ref{tab:q2})}: no feasibility predicate and no rollout, relying entirely on per-step commit-time verification (\emph{seq/none}); a reachability gate without rollout (\emph{reach/none}); a reachability gate with look-ahead (\emph{reach/rollout}); and finally our full method, the learned collision and stability heads paired with rollout (\emph{wm/rollout}).

\setlength{\intextsep}{2pt}
\begin{wraptable}{r}{0.48\linewidth}
\vspace{-0.3cm}
\centering
\caption{\textbf{Predicate ladder} \revise{(group\_max${=}5$) on the held-out test set, timeouts as failures. McNemar is against the row above; \emph{oracle/none} is a depth-0 exact-oracle reference.}}
\label{tab:q2}
\small
\setlength{\tabcolsep}{3pt}
\resizebox{\linewidth}{!}{%
\begin{tabular}{lccc}
\toprule
Predicate & Succ. & McNemar & Orc/br. \\
\midrule
seq/none            & 0.789 & ---                     & 2.07 \\
reach/none          & 0.789 & 0:0, $p{>}0.05$         & 2.07 \\
reach/rollout       & 0.825 & 235:408, $p{<}10^{-10}$ & 0.57 \\
\textbf{wm/rollout (ours)} & \textbf{0.905} & 129:511, $p{<}10^{-50}$ & 0.52 \\
\midrule
oracle/none         & 0.857 & 314:83, $p{<}10^{-30}$  & 1.93 \\
\bottomrule
\end{tabular}%
}
\end{wraptable}

As the Succ.\ and McNemar $p$-values \revise{in Table~\ref{tab:q2}} show, our method succeeds on most structures, and the improvement is significant compared to reach/rollout, indicating that adding look-ahead and the learned predicates each gives a significant paired improvement. Our method also requires the fewest verifier calls per brick (Orc/br.), showing that its brick selection is accurate and relies little on backtracking to recover. For reference we also report an exact-oracle predicate (\emph{oracle/none}), whose verdicts are exact but which cannot afford
look-ahead: rolling it forward times out on 31 of 75 structures (\emph{Exact-oracle+look.} row \revise{of Table~\ref{tab:baselines}}, worst case $5{,}487$s against a $1{,}800$s budget), the very cost the world model removes.

Finally, we compare against external methods under a same $1{,}800$s budget \revise{(Table~\ref{tab:baselines})}, spanning world-model baselines (H-WM~\citep{hwm}, PhysWorld~\citep{physworld}), an optimization baseline (an OR constraint solver~\citep{culbertson}), and two planner-based methods, ASAP~\citep{asap} and WorkBenchMark~\citep{workbenchmark}. All compared methods are re-implemented following their papers or released code and, where trainable, retrained on the StableLego train split.
We use the same structures, the same embodiment, and the same exact static-stability ruler, with our grouping ability switched off ($\mathrm{group\_max}{=}1$) for the planner-based baselines that assume the single-brick action space, while the world-model and optimization baselines retain their own grouping behavior. When compared to planner-based baselines, our method significantly outperforms WorkBenchMark and ASAP (paired McNemar). Our advantage within this shared single-brick action space comes from the planner itself; we also report the result of our method with grouping ability turned on for reference, which solves all structures on the 75-small-structure test set. Against the world-model baselines, our method is stronger while issuing no more exact checks. PhysWorld drives its exact-oracle calls down to our level but does not learn physical feasibility. H-WM, an LLM-scale logical world model, achieves a low success rate as it keeps on hitting the $1{,}800$ time budget. When given a 4-times relaxed budget of $7{,}200$s, it trails our success on the stratified set (achieved 985/1200) while being an order of magnitude slower per step. Against the OR constraint solver, our advantage widens with structure size: the solver must enumerate the whole grouped action space up front, whereas our rollout scores candidates as they arise. Together, these baselines show that generic latent rollouts and up-front optimization do not by themselves preserve physical validity, the gap our learned collision and stability predicates are designed to close. Full re-implementation and training details for all baselines, and the OR formulation, are given in the supplementary material. Some qualitative results of our approach are shown in Figure~\ref{fig:gallery}.

\begin{table}[t]
\centering
\caption{\revise{\textbf{Comparison with baselines.} Succ.\ counts timeouts as failures; McNemar is against Ours on the same set. \textbf{Left:} planner-based baselines (single-brick action space) against Ours (group\_max${=}1$) on the $75$-structure small set ($9$--$32$ bricks). \textbf{Right:} world-model and optimization baselines (own grouping) against Ours (group\_max${=}5$) on the $1200$-structure set stratified by brick count.}}
\label{tab:baselines}
\small
\setlength{\tabcolsep}{3pt}
\resizebox{\textwidth}{!}{%
\begin{tabular}{@{} l c c @{\hspace{1.6em}} l c c @{}}
\toprule
\multicolumn{3}{c}{\revise{\textsc{75-structure small set}}} &
\multicolumn{3}{c}{\revise{\textsc{1200-structure set}}} \\
\cmidrule(r{0.8em}){1-3}\cmidrule(l{0.8em}){4-6}
Method & Succ. ($k/n$) & McNemar &
Method & Succ. ($k/n$) & McNemar \\
\midrule
WorkBenchMark~\citep{workbenchmark} & 0.480 (36/75) & 13:0, $p{<}10^{-3}$ & H-WM~\citep{hwm} & 0.343 (411/1200) & 691:10, $p{<}10^{-144}$ \\
ASAP (GNN)~\citep{asap} & 0.347 (26/75) & 23:0, $p{<}10^{-6}$ & OR solver~\citep{culbertson} & 0.312 (374/1200) & 720:2, $p{<}10^{-156}$ \\
Exact-oracle + look. & 0.587 (44/75) & --- & PhysWorld~\citep{physworld} & 0.809 (971/1200) & 157:36, $p{<}10^{-17}$ \\
Ours (group\_max=1) & 0.653 (49/75) & --- & \revise{\textbf{Ours (group\_max=5)}} & \revise{\textbf{0.910 (1092/1200)}} & \revise{---} \\
\textbf{Ours (group\_max=5)} & \textbf{1.000 (75/75)} & --- & & & \\
\bottomrule
\end{tabular}%
}
\vspace{-0.4cm}
\end{table}

% =====================================================================
\vspace{-0.3cm}
\subsection{Collision Head Analysis}
\label{sec:exp-feas}
\vspace{-0.3cm}
\setlength{\intextsep}{2pt}
\begin{wraptable}{r}{0.48\linewidth}
\vspace{-0.3cm}
\centering
\caption{\textbf{Geometry visibility of the collision head.} FP@R99 (lower is better)
for the two manipulation motions.}
\label{tab:q3}
\small
\begin{tabular}{lccc}
\toprule
Motion & None & Local & Full (ours) \\
\midrule
top-down  & 0.768 & 0.303 & \textbf{0.081} \\
bottom-up & 0.453 & 0.178 & \textbf{0.053} \\
\bottomrule
\end{tabular}
\end{wraptable}

We ask how much of the assembled geometry the head must read. We compare three configurations that share the architecture, the labels, and the training budget, and differ only in the amount of anchor latents input: no anchors at all (\emph{None}, a floor), only anchors within three cells of the target in the $L_\infty$ norm (\emph{Local}), or all of the anchors (\emph{Full}). Beyond this, every configuration still receives the brick type, the target pose, and the scalar scale features. We show the results of two motions separately (top-down placement and bottom-up insertion). As Table~\ref{tab:q3} shows, withholding geometry stops the head from filtering: FP@R99 rises sharply for both motions. Restricting attention to a local neighborhood is also $3.4$--$3.8\times$ worse than the full set. Distant anchors are therefore signal: genuine collisions occur far from the target cell, and a local view cannot anticipate them.

% \begin{table}[t]
% \centering
% \caption{\textbf{Geometry visibility of the collision head.} FP@R99 (lower is better)
% for the two manipulation motions.}
% \label{tab:q3}
% \small
% \begin{tabular}{lccc}
% \toprule
% Motion & None & Local & Full (ours) \\
% \midrule
% top-down  & 0.768 & 0.303 & \textbf{0.081} \\
% bottom-up & 0.453 & 0.178 & \textbf{0.053} \\
% \bottomrule
% \end{tabular}
% \end{table}
% \begin{center}
% \small
% \captionof{table}{\textbf{Geometry visibility of the collision head.}
% FP@R99 (lower is better) for the two manipulation motions.}
% \label{tab:q3}

% \begin{tabular}{lccc}
% \toprule
% Motion & None & Local & Full (ours) \\
% \midrule
% top-down  & 0.768 & 0.303 & \textbf{0.081} \\
% bottom-up & 0.453 & 0.178 & \textbf{0.053} \\
% \bottomrule
% \end{tabular}
% \end{center}

\vspace{-0.4cm}
\section{Conclusion}
\vspace{-0.3cm}

We presented a manipulation-level feasibility world model for brick assembly sequence planning. The model learns the geometry-dependent collision feasibility of a placement over a set of anchor-token latents and rolls them forward with a latent dynamics model, letting the planner look ahead over multiple assembly-by-disassembly steps without querying a physical oracle inside the look-ahead search. On StableText2Lego, this yields higher buildable-assembly success than greedy and heuristic-ordering baselines and stronger results than external assembly-sequence methods under a shared exact stability ruler, while issuing far fewer commit-time oracle calls. Our study has some limitations. The pipeline is validated in simulation with analytic verifiers, leaving real-robot execution under sensing noise and calibration to future work. We treat a coupled brick group as a rigid body that can be grasped and placed as one, without verifying intra-group assembly feasibility. Finally, we model static equilibrium but not the dynamic contact forces that arise during insertion and pressing. Closing these gaps is a natural next step toward physical deployment. 

\subsection*{AI use statement}

We used generative AI tools for the following tasks in this work: implementing
parts of the method code; writing testing and evaluation code and monitoring the
status of training runs; designing and running several of the architecture and
capacity ablations reported in the supplementary material; creating and editing
figures; drafting and editing the text; and summarizing related literature.

We did not use generative AI tools to interpret experimental results, to clean or
reformat datasets, to generate synthetic data, or to formulate or prove
mathematical claims.

We have reviewed all AI-assisted work: AI-generated code was tested by the
authors, AI-assisted experimental results were checked against the exact oracles
and ground-truth labels used throughout the paper, and AI-drafted text was
reviewed and edited by the authors. We take responsibility for the final content
of this work.

\bibliography{iclr2027_conference}

@inproceedings{s2a,
  author       = {Chang Shu and
                  Anton Kim and
                  Shinkyu Park},
  title        = {Subassembly to Full Assembly: Effective Assembly Sequence Planning
                  Through Graph-Based Reinforcement Learning},
  booktitle    = {{IEEE} International Conference on Robotics and Automation, {ICRA}
                  2025, Atlanta, GA, USA, May 19-23, 2025},
  pages        = {14946--14952},
  publisher    = {{IEEE}},
  year         = {2025},
  doi          = {10.1109/ICRA55743.2025.11127710}
}

@inproceedings{grace,
  author       = {Matan Atad and
                  Jianxiang Feng and
                  Ismael Rodr{\'{\i}}guez and
                  Maximilian Durner and
                  Rudolph Triebel},
  title        = {Efficient and Feasible Robotic Assembly Sequence Planning via Graph
                  Representation Learning},
  booktitle    = {{IROS}},
  pages        = {8262--8269},
  year         = {2023},
  doi          = {10.1109/IROS55552.2023.10342352}
}

@inproceedings{asap,
  author       = {Yunsheng Tian and
                  Karl D. D. Willis and
                  Bassel Al Omari and
                  Jieliang Luo and
                  Pingchuan Ma and
                  Yichen Li and
                  Farhad Javid and
                  Edward Gu and
                  Joshua Jacob and
                  Shinjiro Sueda and
                  Hui Li and
                  Sachin Chitta and
                  Wojciech Matusik},
  title        = {{ASAP:} Automated Sequence Planning for Complex Robotic Assembly with
                  Physical Feasibility},
  booktitle    = {{IEEE} International Conference on Robotics and Automation, {ICRA}
                  2024, Yokohama, Japan, May 13-17, 2024},
  pages        = {4380--4386},
  publisher    = {{IEEE}},
  year         = {2024},
  doi          = {10.1109/ICRA57147.2024.10611595}
}

@article{paca,
  author       = {Ruixuan Liu and
                  Alan Chen and
                  Weiye Zhao and
                  Changliu Liu},
  title        = {Physics-Aware Combinatorial Assembly Sequence Planning Using Data-Free
                  Action Masking},
  journal      = {{IEEE} Robotics Autom. Lett.},
  volume       = {10},
  number       = {5},
  pages        = {4882--4889},
  year         = {2025},
  doi          = {10.1109/LRA.2025.3555074}
}

@article{assemblycomplete,
  author       = {Alan Chen and
                  Changliu Liu},
  title        = {AssemblyComplete: 3D Combinatorial Construction with Deep Reinforcement
                  Learning},
  journal      = {CoRR},
  volume       = {abs/2410.15469},
  year         = {2024},
  doi          = {10.48550/ARXIV.2410.15469},
  eprinttype   = {arXiv},
  eprint       = {2410.15469}
}

@inproceedings{brickbybrick,
  author       = {Hyunsoo Chung and
                  Jungtaek Kim and
                  Boris Knyazev and
                  Jinhwi Lee and
                  Graham W. Taylor and
                  Jaesik Park and
                  Minsu Cho},
  editor       = {Marc'Aurelio Ranzato and
                  Alina Beygelzimer and
                  Yann N. Dauphin and
                  Percy Liang and
                  Jennifer Wortman Vaughan},
  title        = {Brick-by-Brick: Combinatorial Construction with Deep Reinforcement
                  Learning},
  booktitle    = {Advances in Neural Information Processing Systems 34: Annual Conference
                  on Neural Information Processing Systems 2021, NeurIPS 2021, December
                  6-14, 2021, virtual},
  pages        = {5745--5757},
  year         = {2021}
}

@inproceedings{icraasp,
  author       = {Lin Ma and
                  Jiangtao Gong and
                  Hao Xu and
                  Hao Chen and
                  Hao Zhao and
                  Wenbing Huang and
                  Guyue Zhou},
  title        = {Planning Assembly Sequence with Graph Transformer},
  booktitle    = {{IEEE} International Conference on Robotics and Automation, {ICRA}
                  2023, London, UK, May 29 - June 2, 2023},
  pages        = {12395--12401},
  publisher    = {{IEEE}},
  year         = {2023},
  doi          = {10.1109/ICRA48891.2023.10160424},
}

@inproceedings{treesba,
  author       = {Mengqi Guo and
                  Chen Li and
                  Yuyang Zhao and
                  Gim Hee Lee},
  editor       = {Ales Leonardis and
                  Elisa Ricci and
                  Stefan Roth and
                  Olga Russakovsky and
                  Torsten Sattler and
                  G{\"{u}}l Varol},
  title        = {TreeSBA: Tree-Transformer for Self-supervised Sequential Brick Assembly},
  booktitle    = {Computer Vision - {ECCV} 2024 - 18th European Conference, Milan, Italy,
                  September 29-October 4, 2024, Proceedings, Part {LXXXVI}},
  series       = {Lecture Notes in Computer Science},
  volume       = {15144},
  pages        = {35--51},
  publisher    = {Springer},
  year         = {2024},
  doi          = {10.1007/978-3-031-73016-0\_3}
}

@article{somacube,
  author       = {Jaehong Oh and
                  Seungjun Jung and
                  Sawoong Kim},
  title        = {Learning to Assemble the Soma Cube with Legal-Action Masked {DQN}
                  and Safe {ZYZ} Regrasp on a Doosan {M0609}},
  journal      = {CoRR},
  volume       = {abs/2508.21272},
  year         = {2025},
  doi          = {10.48550/ARXIV.2508.21272},
  eprinttype   = {arXiv},
  eprint       = {2508.21272}
}

@inproceedings{orasp,
  author       = {Kartik Nagpal and
                  Negar Mehr},
  title        = {Optimal Robotic Assembly Sequence Planning {(ORASP):} {A} Sequential
                  Decision-Making Approach},
  booktitle    = {{IEEE/RSJ} International Conference on Intelligent Robots and Systems,
                  {IROS} 2024, Abu Dhabi, United Arab Emirates, October 14-18, 2024},
  pages        = {9847--9854},
  publisher    = {{IEEE}},
  year         = {2024},
  doi          = {10.1109/IROS58592.2024.10802475}
}

@article{learningtobuild,
  author       = {Jingwen Wang and
                  Johannes Kirschner and
                  Paul Rolland and
                  Luis Salamanca and
                  Stefana Parascho},
  title        = {Learning to Build: Autonomous Robotic Assembly of Stable Structures
                  Without Predefined Plans},
  journal      = {CoRR},
  volume       = {abs/2602.23934},
  year         = {2026},
  doi          = {10.48550/ARXIV.2602.23934},
  eprinttype   = {arXiv},
  eprint       = {2602.23934}
}

@inproceedings{multilevel,
  author       = {Xinghao Zhu and
                  Devesh K. Jha and
                  Diego Romeres and
                  Lingfeng Sun and
                  Masayoshi Tomizuka and
                  Anoop Cherian},
  title        = {Multi-level Reasoning for Robotic Assembly: From Sequence Inference
                  to Contact Selection},
  booktitle    = {{IEEE} International Conference on Robotics and Automation, {ICRA}
                  2024, Yokohama, Japan, May 13-17, 2024},
  pages        = {816--823},
  publisher    = {{IEEE}},
  year         = {2024},
  doi          = {10.1109/ICRA57147.2024.10611259}
}

@article{euphoria,
  author       = {Shih{-}Yu Lai and
                  Chia{-}Ching Yen and
                  Yang{-}Ting Shen and
                  Peter Yichen Chen and
                  Yu{-}Lun Liu and
                  Bing{-}Yu Chen},
  title        = {{EUPHORIA:} Efficient Universal Planning via Hybrid Optimization for
                  Robust Industrial Robotic Assembly},
  journal      = {CoRR},
  volume       = {abs/2605.18872},
  year         = {2026},
  doi          = {10.48550/ARXIV.2605.18872},
  eprinttype   = {arXiv},
  eprint       = {2605.18872}
}

@INPROCEEDINGS{frombricks,
  author={Barghi, Arman and Pasiar, Navid and Masouleh, Mehdi Tale and Kalhor, Ahmad},
  booktitle={2024 10th International Conference on Control, Instrumentation and Automation (ICCIA)}, 
  title={From Bricks to Bots: Automated Collision-Aware Sequence Planning for LEGO Reconstruction with a Two-Finger Gripper}, 
  year={2024},
  volume={},
  number={},
  pages={1-6},
  doi={10.1109/ICCIA65044.2024.10768173}
}

@article{workbenchmark,
  author       = {Wenbo Ma and
                  Daniel Swoboda and
                  Matteo Tschesche and
                  Till Hofmann},
  title        = {WorkBenchMark: {A} LEGO-Based Assembly Benchmark with an Assembly-by-Disassembly
                  Baseline for the Smart Manufacturing League},
  journal      = {CoRR},
  volume       = {abs/2606.19358},
  year         = {2026},
  doi          = {10.48550/ARXIV.2606.19358},
  eprinttype   = {arXiv},
  eprint       = {2606.19358}
}

@inproceedings{grn,
  author       = {Smail Ait Bouhsain and
                  Rachid Alami and
                  Thierry Sim{\'{e}}on},
  title        = {Learning Geometric Reasoning Networks For Robot Task And Motion Planning},
  booktitle    = {The Thirteenth International Conference on Learning Representations,
                  {ICLR} 2025, Singapore, April 24-28, 2025},
  publisher    = {OpenReview.net},
  year         = {2025}
}

@article{fastron,
  author       = {Nikhil Das and
                  Michael C. Yip},
  title        = {Learning-Based Proxy Collision Detection for Robot Motion Planning
                  Applications},
  journal      = {{IEEE} Trans. Robotics},
  volume       = {36},
  number       = {4},
  pages        = {1096--1114},
  year         = {2020},
  doi          = {10.1109/TRO.2020.2974094}
}

@inproceedings{clearancenet,
  author       = {J. Chase Kew and
                  Brian Ichter and
                  Maryam Bandari and
                  Tsang{-}Wei Edward Lee and
                  Aleksandra Faust},
  editor       = {Steven M. LaValle and
                  Ming Lin and
                  Timo Ojala and
                  Dylan A. Shell and
                  Jingjin Yu},
  title        = {Neural Collision Clearance Estimator for Batched Motion Planning},
  booktitle    = {Algorithmic Foundations of Robotics XIV, Proceedings of the Fourteenth
                  Workshop on the Algorithmic Foundations of Robotics, {WAFR} 2021,
                  Oulu, Finland, June 21-23, 2021},
  series       = {Springer Proceedings in Advanced Robotics},
  volume       = {17},
  pages        = {73--89},
  publisher    = {Springer},
  year         = {2021},
  doi          = {10.1007/978-3-030-66723-8\_5}
}

@article{worldmodels,
  author       = {David Ha and
                  J{\"{u}}rgen Schmidhuber},
  title        = {World Models},
  journal      = {CoRR},
  volume       = {abs/1803.10122},
  year         = {2018},
  eprinttype   = {arXiv},
  eprint       = {1803.10122}
}

@inproceedings{interactionnet,
  author       = {Peter W. Battaglia and
                  Razvan Pascanu and
                  Matthew Lai and
                  Danilo Jimenez Rezende and
                  Koray Kavukcuoglu},
  editor       = {Daniel D. Lee and
                  Masashi Sugiyama and
                  Ulrike von Luxburg and
                  Isabelle Guyon and
                  Roman Garnett},
  title        = {Interaction Networks for Learning about Objects, Relations and Physics},
  booktitle    = {Advances in Neural Information Processing Systems 29: Annual Conference
                  on Neural Information Processing Systems 2016, December 5-10, 2016,
                  Barcelona, Spain},
  pages        = {4502--4510},
  year         = {2016}
}

@inproceedings{gns,
  author       = {Alvaro Sanchez{-}Gonzalez and
                  Jonathan Godwin and
                  Tobias Pfaff and
                  Rex Ying and
                  Jure Leskovec and
                  Peter W. Battaglia},
  title        = {Learning to Simulate Complex Physics with Graph Networks},
  booktitle    = {Proceedings of the 37th International Conference on Machine Learning,
                  {ICML} 2020, 13-18 July 2020, Virtual Event},
  series       = {Proceedings of Machine Learning Research},
  volume       = {119},
  pages        = {8459--8468},
  publisher    = {{PMLR}},
  year         = {2020}
}

@inproceedings{planet,
  author       = {Danijar Hafner and
                  Timothy P. Lillicrap and
                  Ian Fischer and
                  Ruben Villegas and
                  David Ha and
                  Honglak Lee and
                  James Davidson},
  editor       = {Kamalika Chaudhuri and
                  Ruslan Salakhutdinov},
  title        = {Learning Latent Dynamics for Planning from Pixels},
  booktitle    = {Proceedings of the 36th International Conference on Machine Learning,
                  {ICML} 2019, 9-15 June 2019, Long Beach, California, {USA}},
  series       = {Proceedings of Machine Learning Research},
  volume       = {97},
  pages        = {2555--2565},
  publisher    = {{PMLR}},
  year         = {2019}
}

@article{dreamerv3,
  author       = {Danijar Hafner and
                  Jurgis Pasukonis and
                  Jimmy Ba and
                  Timothy P. Lillicrap},
  title        = {Mastering Diverse Domains through World Models},
  journal      = {CoRR},
  volume       = {abs/2301.04104},
  year         = {2023},
  doi          = {10.48550/ARXIV.2301.04104},
  eprinttype   = {arXiv},
  eprint       = {2301.04104}
}

@inproceedings{tdmpc2,
  author       = {Nicklas Hansen and
                  Hao Su and
                  Xiaolong Wang},
  title        = {{TD-MPC2:} Scalable, Robust World Models for Continuous Control},
  booktitle    = {The Twelfth International Conference on Learning Representations,
                  {ICLR} 2024, Vienna, Austria, May 7-11, 2024},
  publisher    = {OpenReview.net},
  year         = {2024}
}

@inproceedings{iris,
  author       = {Vincent Micheli and
                  Eloi Alonso and
                  Fran{\c{c}}ois Fleuret},
  title        = {Transformers are Sample-Efficient World Models},
  booktitle    = {The Eleventh International Conference on Learning Representations,
                  {ICLR} 2023, Kigali, Rwanda, May 1-5, 2023},
  publisher    = {OpenReview.net},
  year         = {2023}
}

@article{jepawm,
  author       = {Basile Terver and
                  Tsung{-}Yen Yang and
                  Jean Ponce and
                  Adrien Bardes and
                  Yann LeCun},
  title        = {What Drives Success in Physical Planning with Joint-Embedding Predictive
                  World Models?},
  journal      = {Trans. Mach. Learn. Res.},
  volume       = {2026},
  year         = {2026}
}

@article{hamiltonian,
  author       = {Sen Cui and
                  Jingheng Ma},
  title        = {Physically Native World Models: {A} Hamiltonian Perspective on Generative
                  World Modeling},
  journal      = {CoRR},
  volume       = {abs/2605.00412},
  year         = {2026},
  doi          = {10.48550/ARXIV.2605.00412},
  eprinttype   = {arXiv},
  eprint       = {2605.00412}
}

@inproceedings{cswm,
  author       = {Thomas N. Kipf and
                  Elise van der Pol and
                  Max Welling},
  title        = {Contrastive Learning of Structured World Models},
  booktitle    = {8th International Conference on Learning Representations, {ICLR} 2020,
                  Addis Ababa, Ethiopia, April 26-30, 2020},
  publisher    = {OpenReview.net},
  year         = {2020}
}

@inproceedings{fiocwm,
  author       = {Fan Feng and
                  Phillip Lippe and
                  Sara Magliacane},
  editor       = {Danielle Belgrave and
                  Cheng Zhang and
                  Laura N. Montoya and
                  Hsuan{-}Tien Lin and
                  Razvan Pascanu and
                  Piotr Koniusz and
                  Marzyeh Ghassemi and
                  Nancy Chen and
                  Iv{\'{a}}n Vladimir Meza Ru{\'{\i}}z and
                  Arturo Loaiza{-}Bonilla},
  title        = {Learning Interactive World Model for Object-Centric Reinforcement
                  Learning},
  booktitle    = {Advances in Neural Information Processing Systems 38: Annual Conference
                  on Neural Information Processing Systems 2025, NeurIPS 2025, San Diego,
                  CA, USA, December 2-7, 2025 / Mexico City, Mexico, November 30 - December
                  5, 2025},
  year         = {2025}
}

@article{objectzero,
  author       = {Rodion Vakhitov and
                  Leonid Ugadiarov and
                  Aleksandr Panov},
  title        = {Object-Centric World Models Meet Monte Carlo Tree Search},
  journal      = {CoRR},
  volume       = {abs/2601.06604},
  year         = {2026},
  doi          = {10.48550/ARXIV.2601.06604},
  eprinttype   = {arXiv},
  eprint       = {2601.06604}
}

@article{gwmsurvey,
  author       = {Jiawei Liu and
                  Senqiao Yang and
                  Mingjun Wang and
                  Yu Wang and
                  Bei Yu},
  title        = {Graph World Models: Concepts, Taxonomy, and Future Directions},
  journal      = {CoRR},
  volume       = {abs/2604.27895},
  year         = {2026},
  doi          = {10.48550/ARXIV.2604.27895},
  eprinttype   = {arXiv},
  eprint       = {2604.27895},
  bibsource    = {dblp computer science bibliography, https://dblp.org}
}

@article{gpc,
  author       = {Han Qi and
                  Haocheng Yin and
                  Aris Zhu and
                  Yilun Du and
                  Heng Yang},
  title        = {Inference-Time Enhancement of Generative Robot Policies via Predictive
                  World Modeling},
  journal      = {{IEEE} Robotics Autom. Lett.},
  volume       = {11},
  number       = {5},
  pages        = {5534--5541},
  year         = {2026},
  url          = {https://doi.org/10.1109/LRA.2026.3673995},
  doi          = {10.1109/LRA.2026.3673995},
  bibsource    = {dblp computer science bibliography, https://dblp.org}
}

@article{wav,
  author       = {Yuejiang Liu and
                  Fan Feng and
                  Lingjing Kong and
                  Weifeng Lu and
                  Jinzhou Tang and
                  Kun Zhang and
                  Kevin Murphy and
                  Chelsea Finn and
                  Yilun Du},
  title        = {World Action Verifier: Self-Improving World Models via Forward-Inverse
                  Asymmetry},
  journal      = {CoRR},
  volume       = {abs/2604.01985},
  year         = {2026},
  url          = {https://doi.org/10.48550/arXiv.2604.01985},
  doi          = {10.48550/ARXIV.2604.01985},
  eprinttype   = {arXiv},
  eprint       = {2604.01985},
  bibsource    = {dblp computer science bibliography, https://dblp.org}
}

@inproceedings{brickgpt,
  author       = {Ava Pun and
                  Kangle Deng and
                  Ruixuan Liu and
                  Deva Ramanan and
                  Changliu Liu and
                  Jun{-}Yan Zhu},
  title        = {Generating Physically Stable and Buildable Brick Structures from Text},
  booktitle    = {{IEEE/CVF} International Conference on Computer Vision, {ICCV} 2025,
                  Honolulu, HI, USA, October 19-25, 2025},
  pages        = {14798--14809},
  publisher    = {{IEEE}},
  year         = {2025},
  url          = {https://doi.org/10.1109/ICCV51701.2025.01373},
  doi          = {10.1109/ICCV51701.2025.01373},
  bibsource    = {dblp computer science bibliography, https://dblp.org}
}

@article{stable,
  author       = {Chenhui Xu and
                  Ziyue Bai and
                  Fuxun Yu and
                  Heng Huang and
                  Jinjun Xiong},
  title        = {Rollback-Free Stable Brick Structures Generation},
  journal      = {CoRR},
  volume       = {abs/2605.06947},
  year         = {2026},
  url          = {https://doi.org/10.48550/arXiv.2605.06947},
  doi          = {10.48550/ARXIV.2605.06947},
  eprinttype   = {arXiv},
  eprint       = {2605.06947},
  bibsource    = {dblp computer science bibliography, https://dblp.org}
}

@article{apexmr,
  author       = {Philip Huang and
                  Ruixuan Liu and
                  Changliu Liu and
                  Jiaoyang Li},
  title        = {{APEX-MR:} Multi-Robot Asynchronous Planning and Execution for Cooperative
                  Assembly},
  journal      = {CoRR},
  volume       = {abs/2503.15836},
  year         = {2025},
  url          = {https://doi.org/10.48550/arXiv.2503.15836},
  doi          = {10.48550/ARXIV.2503.15836},
  eprinttype   = {arXiv},
  eprint       = {2503.15836},
  bibsource    = {dblp computer science bibliography, https://dblp.org}
}

@article{vampmr,
  author       = {Philip Huang and
                  Chenrui Gao and
                  Jiaoyang Li},
  title        = {{VAMP-MR:} Vector-Accelerated Motion Planning and Execution for Multi-Robot-Arms},
  journal      = {CoRR},
  volume       = {abs/2607.13478},
  year         = {2026},
  url          = {https://doi.org/10.48550/arXiv.2607.13478},
  doi          = {10.48550/ARXIV.2607.13478},
  eprinttype   = {arXiv},
  eprint       = {2607.13478},
  bibsource    = {dblp computer science bibliography, https://dblp.org}
}

@inproceedings{pointr,
  author       = {Xumin Yu and
                  Yongming Rao and
                  Ziyi Wang and
                  Zuyan Liu and
                  Jiwen Lu and
                  Jie Zhou},
  title        = {PoinTr: Diverse Point Cloud Completion with Geometry-Aware Transformers},
  booktitle    = {2021 {IEEE/CVF} International Conference on Computer Vision, {ICCV}
                  2021, Montreal, QC, Canada, October 10-17, 2021},
  pages        = {12478--12487},
  publisher    = {{IEEE}},
  year         = {2021},
  url          = {https://doi.org/10.1109/ICCV48922.2021.01227},
  doi          = {10.1109/ICCV48922.2021.01227},
  bibsource    = {dblp computer science bibliography, https://dblp.org}
}

@inproceedings{diffpool,
  author       = {Zhitao Ying and
                  Jiaxuan You and
                  Christopher Morris and
                  Xiang Ren and
                  William L. Hamilton and
                  Jure Leskovec},
  editor       = {Samy Bengio and
                  Hanna M. Wallach and
                  Hugo Larochelle and
                  Kristen Grauman and
                  Nicol{\`{o}} Cesa{-}Bianchi and
                  Roman Garnett},
  title        = {Hierarchical Graph Representation Learning with Differentiable Pooling},
  booktitle    = {Advances in Neural Information Processing Systems 31: Annual Conference
                  on Neural Information Processing Systems 2018, NeurIPS 2018, December
                  3-8, 2018, Montr{\'{e}}al, Canada},
  pages        = {4805--4815},
  year         = {2018},
  url          = {https://proceedings.neurips.cc/paper/2018/hash/e77dbaf6759253c7c6d0efc5690369c7-Abstract.html},
  bibsource    = {dblp computer science bibliography, https://dblp.org}
}

@inproceedings{hgt,
  author       = {Ziniu Hu and
                  Yuxiao Dong and
                  Kuansan Wang and
                  Yizhou Sun},
  editor       = {Yennun Huang and
                  Irwin King and
                  Tie{-}Yan Liu and
                  Maarten van Steen},
  title        = {Heterogeneous Graph Transformer},
  booktitle    = {{WWW} '20: The Web Conference 2020, Taipei, Taiwan, April 20-24, 2020},
  pages        = {2704--2710},
  publisher    = {{ACM} / {IW3C2}},
  year         = {2020},
  url          = {https://doi.org/10.1145/3366423.3380027},
  doi          = {10.1145/3366423.3380027},
  bibsource    = {dblp computer science bibliography, https://dblp.org}
}

@inproceedings{visualstability,
  author       = {Wenbin Li and
                  Ales Leonardis and
                  Mario Fritz},
  title        = {Visual Stability Prediction and Its Application to Manipulation},
  booktitle    = {2017 {AAAI} Spring Symposia, Stanford University, Palo Alto, California,
                  USA, March 27-29, 2017},
  publisher    = {{AAAI} Press},
  year         = {2017},
  url          = {http://aaai.org/ocs/index.php/SSS/SSS17/paper/view/15263},
  bibsource    = {dblp computer science bibliography, https://dblp.org}
}

@article{gnnstress,
  author       = {Yuecheng Cai and
                  Jasmin Jelovica},
  title        = {Graph Neural Network for Stress Predictions in Stiffened Panels Under
                  Uniform Loading},
  journal      = {CoRR},
  volume       = {abs/2309.13022},
  year         = {2023},
  url          = {https://doi.org/10.48550/arXiv.2309.13022},
  doi          = {10.48550/ARXIV.2309.13022},
  eprinttype   = {arXiv},
  eprint       = {2309.13022},
  bibsource    = {dblp computer science bibliography, https://dblp.org}
}

@article{gnndisp,
  author       = {Hung{-}Fu Chang and
                  Tzu{-}Kang Lin and
                  Yung{-}Li Cheng},
  title        = {{GNN} for Structural Displacement Prediction},
  journal      = {CoRR},
  volume       = {abs/2605.08303},
  year         = {2026},
  url          = {https://doi.org/10.48550/arXiv.2605.08303},
  doi          = {10.48550/ARXIV.2605.08303},
  eprinttype   = {arXiv},
  eprint       = {2605.08303},
  bibsource    = {dblp computer science bibliography, https://dblp.org}
}

@article{stablelego,
  author       = {Ruixuan Liu and
                  Kangle Deng and
                  Ziwei Wang and
                  Changliu Liu},
  title        = {StableLego: Stability Analysis of Block Stacking Assembly},
  journal      = {{IEEE} Robotics Autom. Lett.},
  volume       = {9},
  number       = {11},
  pages        = {9383--9390},
  year         = {2024},
  doi          = {10.1109/LRA.2024.3414281}
}

@Manual{highs,
    title = {highs: 'HiGHS' Optimization Solver},
    author = {Florian Schwendinger and Balasubramanian Narasimhan and
      Dirk Schumacher},
    year = {2026},
    note = {R package version 1.14.0-2},
    doi = {10.32614/CRAN.package.highs},
  }

@misc{gurobi,
  author       = {{Gurobi Optimization, LLC}},
  title        = {Gurobi Optimizer Reference Manual},
  year         = {2026},
  howpublished = {\url{https://www.gurobi.com}}
}

@article{hwm,
  title   = {Robotic Task and Motion Planning Guided by Hierarchical World Model},
  author  = {Huang, Jinbang and Chen, Wenyuan and Li, Zhiyuan and Pang, Oscar and Hu, Xiao and Zhang, Lingfeng and Hu, Yuanzhao and Coates, Mark and Cao, Tongtong and Quan, Xingyue and Zhang, Zhanguang and Zhang, Yingxue},
  journal = {arXiv preprint arXiv:2602.11291},
  year    = {2026},
}

@article{physworld,
  title   = {Robot Learning from a Physical World Model},
  author  = {Mao, Jiageng and He, Sicheng and Wu, Hao-Ning and You, Yang and Sun, Shuyang and Wang, Zhicheng and Bao, Yanan and Chen, Huizhong and Guibas, Leonidas and Guizilini, Vitor},
  journal = {arXiv preprint arXiv:2511.07416},
  year    = {2025},
}

@inproceedings{culbertson,
  title     = {Multi-Robot Assembly Sequencing via Discrete Optimization},
  author    = {Culbertson, Preston and Bandyopadhyay, Saptarshi and Schwager, Mac},
  booktitle = {IROS},
  year      = {2019},
}

@inproceedings{statebased,
  title     = {State-Based Disassembly Planning},
  author    = {Lei, Chao and Lipovetzky, Nir and Ehinger, Krista A.},
  booktitle = {AAAI},
  year      = {2025},
}
\bibliographystyle{iclr2027_conference}

\appendix
\newpage

% =====================================================================
\section{Experimental Setup and Compute}
\label{app:setup}

\paragraph{Software and hardware.} All learned components use PyTorch $2.4.1$ (CUDA $12.1$) under Python $3.9$, with NumPy $2.0.2$, SciPy $1.13.1$ and pandas $2.3.3$. All training, label generation and planning run on machines with a single NVIDIA L40S GPU ($45$\,GB) and up to $96$ physical CPU cores ($192$ vCPUs).

\paragraph{Training cost.} Every learned component is small and trains with a single seed (Table~\ref{tab:train-cost}). The three components the planner loads at inference time together train in $2.84$ GPU-hours; the stability head is the fourth, and its wall-clock was not recorded. Model-selection sweeps and label generation are excluded.

\begin{table}[htbp]
\centering
\caption{Training cost on a single NVIDIA L40S, single seed. The decoder is trained alongside the encoder and its cost is inside that row. The deployed collision head reads the anchor set pre-embedded over the target structure (Section~\ref{app:abl-head}).}
\label{tab:train-cost}
\footnotesize
\setlength{\tabcolsep}{4pt}
\begin{tabular}{@{}llrll@{}}
\toprule
Component & At inference & Params & Training data & Wall-clock \\
\midrule
Encoder                & yes & $0.926$M    & $3.74$M states, $3$ epochs        & $66$ min \\
Latent dynamics        & yes & $42.4$k     & $3.69$M removals, both directions & $60$ min \\
Collision head         & yes & $106{,}518$ & $6.47$M cells, $40$ epochs        & $44.5$ min \\
Stability head         & yes & $1.20$M     & $21{,}093$ states                 & not recorded \\
\bottomrule
\end{tabular}
\end{table}

\paragraph{Label-generation cost.} Producing the exact labels costs more than training the models that consume them, which is the data-side form of our premise that the exact oracle is expensive. For the collision head, one label records whether one gripper configuration can move one candidate brick or group onto one partial structure without collision. We generated $10.51$M such labels over $416{,}493$ partial states in $16.7$ CPU-hours. The $29.6\%$ whose failure cause is an unreachable inverse-kinematics solution are not collision labels and are handled by the enumerated gate of Section~\ref{app:abl-gate}; what remains is the head's training set of $6.47$M cells, $65.3\%$ of them collision-free, and actions moving a group of two or more bricks are $40.6\%$ of the $360{,}061$ training states. For the stability head, one label records whether one partial structure stands under gravity according to the exact program of Section~\ref{app:lp}; generating the $19{,}993$ structures we sample ourselves costs $4$ further CPU-hours on $4$ cores. Against these $20.7$ CPU-hours of labelling stands $2.84$ GPU-hours of training.

% =====================================================================
\section{Parameterization and Implementation Details}
\label{app:params}

This appendix gives the two quantities the main paper defers, the per-brick descriptor and the discrete action expansion, then the implementation constants.

\subsection{The per-brick descriptor}
\label{app:params-feat}

Every module summarizes a brick by the same eleven numbers (Table~\ref{tab:brick-feat}) rather than by its full geometry. Three entries deserve comment. The mass comes from a table calibrated against real LEGO bricks and is therefore not proportional to the stud count: a $2{\times}2$ and a $1{\times}4$ both cover four studs but weigh $2.67$ and $3.65$ units. Dimension $4$ marks bricks whose short side is a single stud, which matters because the stability model of Section~\ref{app:lp} attaches four contact points to each stud of such a brick and three to each stud of a wider one. Dimension $10$ carries no information and is kept only so the width matches the one used when the labels were generated.

\begin{table}[htbp]
\centering
\caption{The eleven-dimensional per-brick descriptor, read by the encoder, the dynamics model and the stability head. Here $d_1,d_2$ are the footprint dimensions in studs and $z$ is the layer the brick sits in.}
\label{tab:brick-feat}
\begin{tabular}{clc}
\toprule
Dim & Meaning & Changes during assembly \\
\midrule
$0$  & footprint length $d_1$ in studs                        & no \\
$1$  & footprint width $d_2$ in studs                         & no \\
$2$  & short side, $\min(d_1,d_2)$                            & no \\
$3$  & footprint area, $d_1 d_2$                              & no \\
$4$  & one stud wide, i.e.\ $\min(d_1,d_2)$ equals one        & no \\
$5$  & mass from a calibrated table, a $1{\times}1$ being $1.0$ & no \\
$6$  & relative height $z/z_{\max}$                           & \textbf{yes} \\
$7$  & number of studs supporting it from below               & \textbf{yes} \\
$8$  & number of studs it supports from above                 & \textbf{yes} \\
$9$  & rests on the ground, i.e.\ $z$ equals zero             & no \\
$10$ & reserved slot, held at $0$                             & --- \\
\bottomrule
\end{tabular}
\end{table}

\subsection{The gripper configurations of one action}
\label{app:params-cells}

An action says which brick or rigidly-coupled group to move and by which of the two motions, but not how the arms carry that out. Expanding the remaining choices gives a fixed set of $112$ ways to execute it, and we call one element a \emph{gripper configuration}: which of two arms performs the motion, which of the four sides of the brick the gripper presses from, which of seven offsets of the pressing tool along that side is used, and which of two directions the tool approaches from, giving $2\times4\times7\times2=112$.

For bottom-up insertion all four choices matter, since the approach direction flips the sign of the lateral offset at which the tool enters. For top-down placement it has no effect: the routine generating that motion takes the index but never uses it, so the $56$ slots carrying one value produce pose chains identical to the $56$ carrying the other. We fix one value as canonical and mask the other $56$ away. Nothing decides this at run time; the mask is a function of the motion type alone, known before any label exists, and it is applied wherever the slots are touched, so the label generator emits labels only for unmasked slots, the training loss skips masked ones, and the planner reads only unmasked ones. The head still emits $112$ numbers so both motion types share one output layer, but for top-down placement the caller discards half.

 A motion is a short chain of keyframe poses rather than a continuous trajectory, and feasibility is decided on that chain: the arm must have an inverse-kinematics solution at every keyframe and the tool and carried bricks must be clear there. Top-down placement has five keyframes, an offset pose below the target, a lift, the target pose, a rotation of the tool that seats the studs, and a final short lift; bottom-up insertion has four, an approach pose, a tilted pose, a descent and a lift. Every pose is generated from the pose of the target stud by offsets that are not free, being either tool-calibration constants or a fixed expression in the brick footprint, and neither generator accepts a height or clearance argument. The vertical geometry is therefore determined once the target stud is fixed, which is what makes these $112$ slots the entire action space of one candidate. 

\subsection{Implementation constants}
\label{app:params-impl}

\paragraph{Encoder and decoder.} An anchor neighborhood is $3^3=27$ cells, each described by its occupancy, six exposure bits and the descriptor of the brick occupying it. The encoder embeds each cell, applies two blocks of local 3D convolution at width $128$, and pools with a mask, giving one latent of width $32$ ($0.926$M parameters). Its training signal comes from a decoder mapping $32\rightarrow256\rightarrow27$ occupancy logits ($15{,}387$ parameters), which is discarded before planning. The encoder is then frozen and every module below trains on top of it. The number of anchors is not a constant: covering one state needs a median of $42$ anchors, while the set the planner uses is chosen once over the target structure and has a median of $73$, a $90$th percentile of $154$ and a maximum of $285$ over $1{,}239$ validation structures, with $39\%$ to $47\%$ of its anchors sitting on cells that are empty at any given moment.

\paragraph{Latent dynamics.} A gated local-residual perceptron with two hidden layers of width $128$ ($42.4$k parameters) whose last layer is zero-initialized, so before training it leaves every latent unchanged. Its action vector is $137=125+11+1$: a dilated $5^3$ occupancy mask of the moved brick, taken one cell wider than the anchor's own neighborhood because a brick just outside can cover the exposed faces of its boundary cells; the moved brick's descriptor; and one direction bit, which is what lets one set of weights serve both directions, since placing a brick and removing it give vectors agreeing in all of the first $136$ entries. Anchors to renew are chosen by dilating the moved brick's cells one step along each of the six axis directions.

\paragraph{Collision head.} A cross-attention transformer of width $64$, three layers, four heads ($106{,}518$ parameters); the $112$ query tokens never attend to one another, so there is no self-attention. A query is the sum of four learned embeddings, one per choice above, and a projection of a $19$-dimensional candidate summary: six numbers for footprint, area and mass of the gripped brick, four for its position and ground contact, one for orientation, three for the size of the structure, five for the size of the moved group. A key is the sum of a projection of the anchor latent from $32$ to $64$, shared by both anchor sets, a projection of the offset from the anchor cell to the brick center, and a tag for the built structure or the moved group, plus one always-valid null token. The attention logits receive a bias from a two-layer network of width $16$ over that offset and its length, one scalar per head, computed once per anchor and zero for the null token. The motion type enters by feature-wise modulation at every layer, the feedforward expansion is two, and the output is $112$ logits and $112$ clearances in millimetres, negative meaning interpenetration. The head is not given the pose of the pressing point or of the tool, the joint angles, the arm link geometry, or the geometry of the table and part feeder.

\paragraph{Gates and stability head.} The press gate is a closed-form test that the stud carrying the assembly force lies within the tool's admissible bounds. The reachability gate is a lookup table over the predicate's finite domain, $11{,}289{,}600$ and $5{,}644{,}800$ slots for the two motions, six minutes to enumerate. The stability head is a message-passing network over the brick adjacency graph: nodes are bricks carrying the descriptor above, each vertically adjacent pair contributes two directed edges carrying the overlapping stud count, that count over the smaller footprint, an upward bit and a ground bit, and the ground is one extra node joined to every brick in the bottom layer. Ten rounds at width $128$ precede a two-layer readout ($1{,}203{,}970$ parameters), and a structure is stable when no brick exceeds $\tau{=}0.85$. 

% =====================================================================
\section{Planning Experiments: Protocol and Caveats}
\label{app:protocol}

\paragraph{Success.} A structure counts as a success only if its recorded sequence replays step by step under both exact references, that is every step passes the exact collision oracle and the exact stability program of Section~\ref{app:lp}, and every brick is removed. A sequence containing any unstable step is a failure even if it runs to completion. This is deliberately not the check the planner performs during search, since whether a planner re-checks a candidate before committing is part of the method being compared; the reported success rate never uses it and comes from one path only, an offline replay applying the same two exact references to every arm.

\paragraph{Statistics.} All intervals are Wilson score intervals at $95\%$ confidence, from the number of successes and the number of evaluable structures. All paired tests are exact binomial McNemar tests, conditioned on the discordant pairs and doubled in the smaller tail, which is conservative. Interval overlap is never used to compare two arms; only the paired test is.

% =====================================================================
\section{Ablation Studies}
\label{app:ablation}

 Every architectural choice was made on the validation split; for the configurations we deploy we also re-report final numbers on held-out test.

\subsection{Anchor sets}
\label{app:abl-repr}

\paragraph{Ablation purpose and setup.} The encoder and its decoder only: the encoder compresses one anchor neighborhood into a latent of width $D$ and the decoder must recover the $27$ occupancy values. The question is how to divide a fixed amount of memory between the number of anchors and the width of each. We score a latent by requiring all $27$ decoded values to match the truth, and call that rate \emph{exact-block} accuracy; the same measure is used for the dynamics model in Section~\ref{app:abl-dyn}. Every such measurement carries a control in which the evaluation is repeated after shuffling the cells of the input, which lands between $0.57$ and $0.66$ for every arm we trained, confirming the measure has not collapsed into predicting a fixed prior. All numbers here are on validation.

\paragraph{Results.} The neighborhood is a cube of side $2h{+}1$, so the radius $h$ controls both how much geometry one anchor must summarize and how many anchors cover the structure, and the two move in opposite directions: at $h{=}1$ a neighborhood holds $27$ cells and a typical state needs a median of $42$ anchors, at $h{=}2$ it holds $125$ and needs $17$, at $h{=}3$ it holds $343$ and needs $9$. The memory spent is the number of anchors times the width, so the fair comparison is at equal values of that product (Table~\ref{tab:abl-kd}).

Within every band the small-neighborhood, many-anchor configuration wins by a factor of $1.5$ to $2$, and at the working point it is already near-lossless, $0.9993$ at width $16$ and $1.0000$ at width $32$, whereas the same memory in a coarser neighborhood falls far short. This is the quantitative reason we spread geometry over a set of local anchor latents rather than compressing it into one wide global vector: at equal capacity, locality buys more than per-token width. The table also locates the width at which reconstruction stops improving, which is $16$, so the deployed width of $32$ is a margin rather than a requirement. On test the deployed encoder fails on $0$ of $998{,}209$ anchors, a Poisson $95\%$ upper bound of $3.7\times10^{-6}$; this is the selected configuration re-scored, not a new ranking. We verified separately that the frozen latent already carries the occupancy of the anchor's own center cell, so the encoder needs no retraining for the pre-embedded anchor set, which places anchors on cells that are currently empty.

\begin{table}[htbp]
\centering
\caption{Exact-block accuracy at matched capacity, capacity being the number of anchors $K$ times the latent width $D$; each entry names the width $D$ placing that radius inside the band. All arms share one decoder family, which at width $8$ can itself move this metric by up to $0.05$.}
\label{tab:abl-kd}
\begin{tabular}{lccc}
\toprule
Capacity $K\!\cdot\!D$ & $h{=}1$ ($27$ cells) & $h{=}2$ ($125$ cells) & $h{=}3$ ($343$ cells) \\
\midrule
$100$--$200$   & ---                 & $0.412$ ($D{=}8$)  & $0.287$ ($D{=}16$) \\
$200$--$400$   & $\mathbf{0.946}$ ($D{=}8$)  & $0.687$ ($D{=}16$) & $0.465$ ($D{=}32$) \\
$400$--$800$   & $\mathbf{0.999}$ ($D{=}16$) & $0.906$ ($D{=}32$) & $0.630$ ($D{=}64$) \\
$800$--$1600$  & $\mathbf{1.000}$ ($D{=}32$) & $0.990$ ($D{=}64$) & --- \\
\bottomrule
\end{tabular}
\end{table}

\subsection{Latent dynamics}
\label{app:abl-dyn}

\paragraph{Ablation purpose.} Input one anchor latent plus one action vector, output the updated latent of that anchor, scored by exact-block accuracy. The denominator is always the set of anchors the update renews, since the others are unchanged by construction and would inflate every number. Arms are on validation; the deployed model, which runs in both directions, reaches $0.99825$ there and $0.99895$ on held-out test.

\paragraph{References.} The ceiling, exact re-encoding, runs the frozen encoder again on the true next state for the anchors that changed; being given the answer, no local update can beat it, and it is not free, since selecting those anchors requires the exposed faces of the next state, the geometric computation a rollout exists to avoid. The floor, carry-forward, applies no update at all. Because the model's last layer is zero-initialized, an untrained model is identical to the floor cell for cell, so the floor is literally the starting point of training.

\paragraph{Results.} Withholding the mask is fatal, $1{,}318$ failures against $8$ (Table~\ref{tab:dyn-enc}). The dilated ring then removes three quarters of what remains, from $8$ to $2$, matching its purpose. At $2$ failures against $1$ for the ceiling the learned update and the ceiling are indistinguishable at this sample size, while the floor fails on $10{,}245$. This scan runs $2{,}500$ optimizer steps, about a tenth of an epoch. Sweeping hidden width over $64$, $128$ and $256$ and depth over $2$ and $3$ leaves the deployed setting already at the ceiling.

\begin{table}[htbp]
\centering
\caption{How the action is described, at matched hidden width $128$ and depth $2$; parameter counts differ only because the first layer follows the action vector. Scored on $16{,}903$ renewed anchors; the lower block is the ceiling and floor defined in the text.}
\label{tab:dyn-enc}
\begin{tabular}{lccc}
\toprule
Action description & Params & Failures & Exact-block \\
\midrule
center offset only ($14$ numbers) & $26.7$k & $1318$  & $0.9220$ \\
inner mask ($38$ numbers)         & $29.7$k & $8$     & $0.9995$ \\
dilated mask ($136$ numbers, ours) & $42.3$k & $\mathbf{2}$ & $\mathbf{0.9999}$ \\
\midrule
exact re-encoding (ceiling)       & ---     & $1$     & $0.9999$ \\
carry-forward (floor)             & ---     & $10245$ & $0.3939$ \\
\bottomrule
\end{tabular}
\end{table}

\subsection{Collision head}
\label{app:abl-head}

\paragraph{Module size.} Sweeping latent width against depth, three layers at width $64$ match two layers at width $96$ with a third fewer parameters, so we use three layers at moderate width; the attention head count has no measurable effect (spread $0.0066$ over one to eight heads), and we fix it at four. The model is trained up to 40 epochs on full training set.

\paragraph{Joint angles of the arm is not in the input.} Appending the inverse-kinematics solution at each keyframe does not help and quietly hurts. Ranking quality is unchanged, $0.9885$ against $0.9879$, but the false-positive rate at the operating threshold is $19\%$ higher with the joint angles, $0.1441$ against $0.1211$, while on a split drawn from the same objects as training it is lower, $0.0990$ against $0.1099$: the signature of overfitting. That is expected, since the joint solution is a deterministic function of inputs the head already has. The experiment is run on bottom-up insertion only, because for top-down placement the label pipeline stops at the first failing keyframe, so the number of recorded joint angles reveals how far the motion got and would leak the label through the length of the input.

\begin{figure}[t]
\centering
\includegraphics[width=\columnwidth]{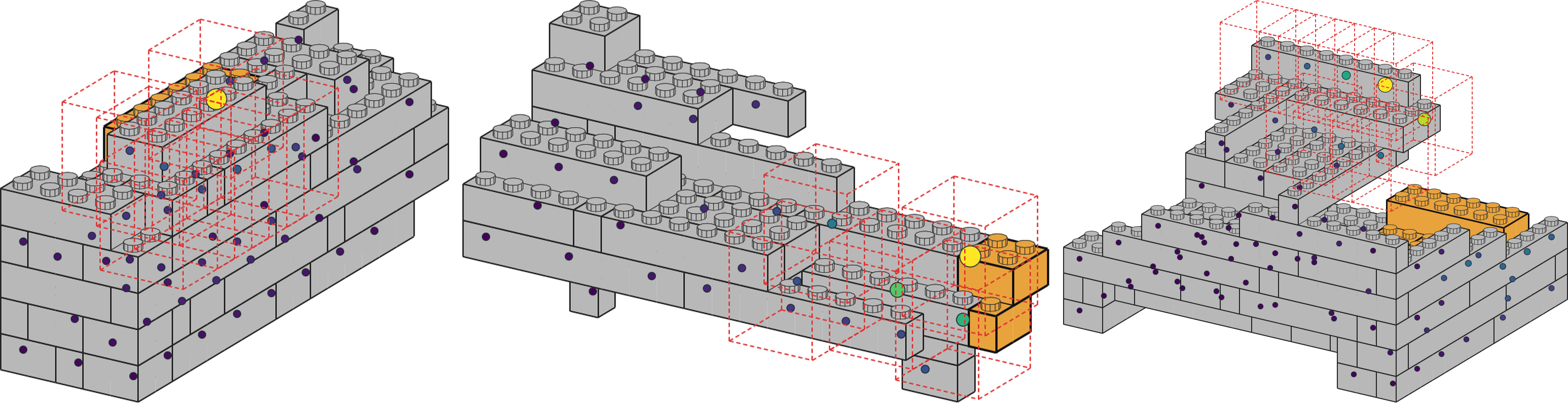}
\caption{The collision head attends to the geometry that actually blocks the motion. Three representative top-down (vertical-placement) cases from the test split, one per group size ($k{=}1,2,4$, left to right). Grey bricks are the built structure; the highlighted bricks are the candidate group whose press point is being judged. Dots are the anchors of the built structure; each dot encodes the attention weight this press point places on that anchor, where \emph{a larger, brighter dot means more attention}.
Red dashed cubes mark the culprit-anchor neighbourhoods covering the brick that the exact
oracle reports as the blocker, and in every
panel the attention collapses onto the anchors that lie inside the red
cubes.}
\label{fig:attn}
\end{figure}

\paragraph{Attention localization.}
We asked whether the trained head attends to the geometry actually responsible for a collision. Using a target computed by the oracle and independent of the head, we record which brick blocks the placement and check whether the head's most-attended anchor covers that brick (Table~\ref{tab:attn-loc}; Figure~\ref{fig:attn} shows one instance). The head localizes the blocker far above chance overall, and almost always for top-down placement and for grouped removals.
Bottom-up insertion is the exception: this action first approaches the built structure horizontally and then lifts the brick to attach, so the colliding part is usually the tool or the wrist rather than the carried brick, and the responsible brick sits far from the target cell (median $L_\infty$ distance $8$ cells, against $5$ for top-down). The low bottom-up number is thus a limit of the per-brick attribution, not of the head's filtering, since the FP@R99 on bottom-up is in fact the lower of the two motions.

\begin{table}[t]
\centering
\caption{\textbf{Attention localization of the collision head.} For each collision cell,
the oracle independently reports which brick blocks the placement; we check whether the
head's most-attended anchor covers that brick. \emph{Top-1} is the fraction of collision
cells for which it does, and \emph{Random} is the chance level of picking one anchor
uniformly. \emph{grouped} denotes removals of two or more rigidly-coupled bricks and
cuts across the two motions rather than adding to them.}
\small
\begin{tabular}{lccc}
\toprule
Subset & $n$ & Top-1 & Random \\
\midrule
Overall            & 1765 & 0.607 & 0.121 \\
\quad top-down     & 906  & 0.940 & 0.117 \\
\quad bottom-up    & 859  & 0.256 & 0.126 \\
grouped ($\ge 2$)  & 152  & 0.921 & 0.107 \\
\bottomrule
\end{tabular}
\label{tab:attn-loc}
\end{table}

\subsection{Reachability gate}
\label{app:abl-gate}

\paragraph{Ablation setup.} The main paper resolves reachability by table lookup rather than by learning; this subsection gives the evidence. The learned version we compare against takes the candidate brick's shape, orientation and target cell and emits the reachable-or-not bit for all of that brick's configurations in one forward pass, with $230$k parameters, more than twice the collision head, trained for $200$ epochs. Its input domain is $14$ combinations of brick shape and orientation times the target cell over a $20\times20$ footprint and $18$ layers, which is $100{,}800$ distinct inputs.

We hold out along three axes, each matching a way deployment could differ from training. \emph{Random} is a uniform split and serves as the memorization control. \emph{Unseen heights} trains on target cells up to layer $14$ and tests on layer $15$ and above, which a planner on taller structures would face. \emph{Unseen brick shapes} holds out two of the $14$ shape and orientation combinations, which adding a brick type would face. \emph{Unseen board region} trains on cells with one planar coordinate at most $29$(x=9~29) and tests beyond, which moving the workpiece would face.

The outcome is the same for both motions (Table~\ref{tab:abl-reach}). On the random split the head reproduces the table almost exactly, so capacity is not the issue. Under every held-out split it collapses: on unseen heights the false-positive rate at $99.9\%$ recall rises to $0.336$ and $0.390$, and on unseen shapes and regions it reaches $1.0$, meaning the threshold must admit every configuration and the gate rejects nothing. A gate that cannot be trusted on a height, shape or region it has not seen has no advantage over enumeration, and a lookup is in addition about three times faster than a forward pass, $0.044\,\mu$s against $0.127\,\mu$s, so we enumerate. This draws the boundary for what the world model should learn at all: a predicate whose domain is finite, structure-independent and enumerable should be tabulated, since learning it buys nothing and introduces a failure mode, and what is left is the collision predicate, whose input includes the assembled geometry.

\begin{table}[htbp]
\centering
\caption{A learned reachability predicate against the enumerated table. FP@R99.9 is the false-positive rate at a recall of $99.9\%$, where $1.0$ means the gate rejects nothing. The random split is a memorization control.}
\label{tab:abl-reach}
\small
\begin{tabular}{lcccc}
\toprule
 & \multicolumn{2}{c}{Top-down placement} & \multicolumn{2}{c}{Bottom-up insertion} \\
\cmidrule(lr){2-3}\cmidrule(lr){4-5}
Split & Accuracy & FP@R99.9 & Accuracy & FP@R99.9 \\
\midrule
random (control)     & $99.96\%$ & $0.0004$ & $99.98\%$ & $0.0000$ \\
unseen heights       & $89.00\%$ & $0.390$  & $96.80\%$ & $0.336$ \\
unseen brick shapes  & $95.43\%$ & $1.0$    & $91.80\%$ & $1.0$ \\
unseen board region  & $78.68\%$ & $1.0$    & $89.55\%$ & $1.0$ \\
\bottomrule
\end{tabular}
\end{table}

\subsection{Stability head}
\label{app:abl-stab}

\paragraph{Ablation setup.} Input the brick adjacency graph of a partial structure, output one instability probability per brick, with a structure stable when no brick exceeds $\tau$. Two things are decided: whether this head can be trained on the labels the corpus provides, and whether it is accurate enough to be the final word on stability. The answers are no and no, and the second is why the planner re-checks stability exactly before committing a step.

\paragraph{Training data generation.} The exact program of Section~\ref{app:lp} marks individual bricks, so one can ask what kind of failure the labels contain. In the complete objects of the corpus, every brick it marks is unsupported: it hangs in the air with nothing beneath it. That is not what a disassembly planner meets. Of the failures occurring in real planning runs, $96\%$ are a single stud connection carrying more force than one stud can hold and the remainder are a sparse remnant tipping under its own torque. Neither appears in the corpus labels.

 We therefore sample partial structures ourselves, from the training split only, and which sampler to use is itself an experiment (Table~\ref{tab:abl-stab-sampler}). A uniform random subset is nearly useless: $94\%$ of its states are unstable but overwhelmingly for the one reason we already have in abundance, a brick left hanging. Requiring the remainder to stay connected to the ground removes that class at a stroke. But ground connectivity alone does not produce tipping; only when the remnant is peeled to between five and twenty bricks does tipping appear at all, and that sampler is also the fastest. A fourth, hollowing a structure into an arch on the expectation that it would tip, fails for a physical reason: the friction of the studs holds the arch up, so its failures are recorded as stud overload.

\begin{table}[htbp]
\centering
\caption{Four samplers for generating partial structures. Unstable is the fraction of generated states the exact program judges unstable; the next three columns break those down by cause. The last row is the sampler we use.}
\label{tab:abl-stab-sampler}
\small
\setlength{\tabcolsep}{4pt}
\begin{tabular}{@{}lcccccc@{}}
\toprule
Sampler & Median bricks & Unstable & Stud overload & Tipping & Hanging brick & s/state \\
\midrule
uniform random subset            & $39$ & $94\%$   & $17$ & $2$ & $75$ & $0.73$ \\
peel, keeping ground contact     & $36$ & $51\%$   & $51$ & $0$ & $0$  & $1.01$ \\
hollow out into an arch          & $77$ & $15\%$   & $21$ & $0$ & $0$  & $2.63$ \\
peel to $5$--$20$ bricks (ours)  & $12$ & $44.5\%$ & $83$ & $6$ & $0$  & $\mathbf{0.20}$ \\
\bottomrule
\end{tabular}
\end{table}

Every generated state is first checked for structural validity, that every brick shape comes from the library, no cell is claimed by two bricks and nothing falls outside the grid. The labels then needed cleaning, because some were artifacts of the solver. Re-checking $31$ unstable validation states under the final protocol leaves $24$ genuinely unstable ones, $23$ stud overload and one tipping; of the seven discarded, four were judged stable even under the original seed, so those labels were simply wrong. The cause is that the criterion declares a brick unbalanced when an equality residual exceeds $10^{-6}$, exactly the solver's default feasibility tolerance, so a state whose residual sits there can be judged either way; the state that flipped had a residual of $1.116\times10^{-6}$. Every stability label in this paper is produced under the corrected protocol of Section~\ref{app:lp}. The result is $19{,}993$ partial structures, of which $10{,}362$ are unstable, $9{,}879$ from stud overload and $483$ from a pure force or torque imbalance. We train on a mixture of these and $1{,}100$ complete objects so the head keeps its accuracy on finished structures; validation is used only to calibrate $\tau$, and the test split is not used at any point in building this head.

\paragraph{Training results.} Joint training raises state-level ranking quality from $0.860$ to $0.973$ and shifts the whole operating curve (Table~\ref{tab:abl-stab-tau}). The comparison to keep in view is the top row: at its calibrated threshold, the surrogate trained only on complete objects misses every one of the $24$ genuinely unstable states. The retrained head at $\tau{=}0.85$ misses three of $24$. 

\begin{table}[htbp]
\centering
\caption{Stability surrogates and thresholds, on $6{,}553$ validation states containing $24$ genuinely unstable ones. Missed unstable counts states the gate wrongly passes; false rejection is the fraction of stable states it wrongly blocks.}
\label{tab:abl-stab-tau}
\small
\begin{tabular}{llcc}
\toprule
Surrogate & $\tau$ & Missed unstable & False rejection \\
\midrule
trained on complete objects only, $5$ rounds & $0.90$ & $24/24$ ($100\%$) & $0.0\%$ \\
trained on complete objects only, $5$ rounds & $0.80$ & $11/24$ ($46\%$)  & $1.2\%$ \\
retrained on partial structures, $10$ rounds & $0.90$ & $6/24$ ($25\%$)   & $1.7\%$ \\
retrained on partial structures, $10$ rounds (deployed) & $0.85$ & $3/24$ ($12\%$) & $2.5\%$ \\
retrained on partial structures, $10$ rounds & $0.80$ & $2/24$ ($8\%$)    & $3.6\%$ \\
\bottomrule
\end{tabular}
\end{table}

% =====================================================================
\section{LP Reformulation of the StableLego Stability Program}
\label{app:lp}

Our exact stability check is a linear program derived from the force-equilibrium analysis of StableLego~\citep{stablelego}. We state the original program, state ours, prove in what sense the two agree, and report the measured agreement. 

\paragraph{Original program.} Original program searches for contact forces that would hold the structure in equilibrium, penalizing how far from equilibrium it must leave each brick, and a brick is declared unstable if the best achievable solution still leaves it out of balance or if holding it requires more force than one stud connection can carry.

Let $\mathcal{B}$ be the bricks. Two vertically adjacent bricks meet at one or more stud cells; write $\mathcal{C}$ for those contacts. At a contact $c$ the model uses not one force but $P_c$ of them, where $P_c$ is four if the brick on that side is one stud wide and three otherwise, which is why dimension $4$ of the per-brick descriptor exists. For each contact $c$ and point $p$ there are four nonnegative variables: $f^{\uparrow}_{cp}$ and $n^{\downarrow}_{cp}$ for the force the lower brick exerts upward and the reaction it receives downward, and $f^{\downarrow}_{cp}$ and $n^{\uparrow}_{cp}$ for the same pair seen from the upper brick. Further nonnegative variables carry the horizontal press between studs in the four lateral directions. For each brick $i$ the program forms five residuals, the net force along each axis, $r^x_i, r^y_i, r^z_i$, and the net torque about each horizontal axis, $\tau^1_i, \tau^2_i$, all linear in the contact variables, with the vertical one carrying the weight $w_i$ and the torques carrying lever arms that are constants of the geometry. The residuals are then made absolute, and $m_i$ collects the largest downward force on any one stud of brick $i$. With $\alpha=10^{-3}$ and $\beta=10^{-6}$,
\begin{equation}
\label{eq:orig}
\begin{aligned}
\min\quad
& \sum_{i\in\mathcal{B}} \big(a^x_i + a^y_i + a^z_i + b^1_i + b^2_i\big)
  \;+\; \alpha \sum_{i\in\mathcal{B}} m_i
  \;+\; \beta \sum_{c,p} f^{\downarrow}_{cp} \\
\text{s.t.}\quad
& r^x_i,\ r^y_i,\ \tau^1_i,\ \tau^2_i \ \text{are the signed sums of the contact terms acting on brick } i, \\
& r^z_i \ \text{is that same sum along the vertical axis, minus the weight } w_i, \\
& f^{\uparrow}_{cp} = f^{\downarrow}_{c'p}, \qquad n^{\downarrow}_{cp} = n^{\uparrow}_{c'p}, \\
& a^x_i = \lvert r^x_i\rvert,\ \ a^y_i = \lvert r^y_i\rvert,\ \ a^z_i = \lvert r^z_i\rvert,\ \
  b^1_i = \lvert \tau^1_i\rvert,\ \ b^2_i = \lvert \tau^2_i\rvert, \\
& m_i = \max\nolimits_{\,c\in i,\,p} f^{\downarrow}_{cp}, \\
& n^{\downarrow}_{cp}\, f^{\uparrow}_{cp} = 0, \qquad n^{\uparrow}_{cp}\, f^{\downarrow}_{cp} = 0,
\end{aligned}
\end{equation}
required for every brick $i\in\mathcal{B}$ and every contact point $c,p$. The third line pairs each contact seen from below with the same contact seen from above, so the two bricks exert equal and opposite forces on each other, and $c'$ denotes that opposite view. All force variables are nonnegative except the net vertical residual, which is free. Given an optimal solution, brick $i$ is unstable when any of its five absolute residuals exceeds a tolerance $\varepsilon = 10^{-6}$, meaning no admissible forces balance it, or when $T - \max_{c\in i,p} f^{\downarrow}_{cp} \le 0$ with $T = 0.98$\,N, meaning some stud would carry more than one stud connection holds. The structure is buildable when no brick is unstable.

 The objective is linear, and the nonlinearity is entirely in the constraints. The last line is a set of bilinear \emph{equalities} forbidding a contact point from carrying a force and its opposing reaction at once. Separately, the absolute-value and maximum constraints are piecewise linear and a solver expands each internally with auxiliary binaries. And no friction cone or second-order cone appears anywhere, since the directional forces are independent nonnegative variables with linear bounds. This original equation is solved by StableLego with a commercial solver~\citep{gurobi}.

\paragraph{Our reformulation.} We obtain a linear program by two exact rewrites and one relaxation, writing $\mathcal{A}$ for the index set of the five absolute values per brick:
\begin{equation}
\label{eq:ours}
\begin{aligned}
\min\quad
& \sum_{i\in\mathcal{B}} \big(a^x_i + a^y_i + a^z_i + b^1_i + b^2_i\big)
  \;+\; \alpha \sum_{i\in\mathcal{B}} m_i
  \;+\; \beta \sum_{c,p} f^{\downarrow}_{cp} \\
\text{s.t.}\quad
& \text{the equilibrium and pairing constraints of~\eqref{eq:orig}, unchanged,} \\
& a_i \ge r_i, \qquad a_i \ge -r_i \qquad \text{for every } a\in\mathcal{A}
  \text{ and its residual } r, \\
& m_i \ge f^{\downarrow}_{cp} \qquad \text{for every contact point of brick } i,
\end{aligned}
\end{equation}
required for every brick $i\in\mathcal{B}$, with the same nonnegativity, the same free vertical residual, and the bilinear equalities removed. Being a pure linear program, it is solved with the open-source solver HiGHS~\citep{highs}, which is the only stability solver our pipeline uses.

The first rewrite replaces $a = \lvert r \rvert$ by the pair $a \ge r$ and $a \ge -r$, and it is exact rather than a relaxation because $a$ enters the minimized objective with coefficient $+1$, so at an optimum one inequality is tight and $a$ equals $\lvert r\rvert$. The second replaces $m_i = \max_p f^{\downarrow}_{cp}$ by $m_i \ge f^{\downarrow}_{cp}$, exact for the same reason since $\alpha > 0$. The third change is the relaxation: we drop the bilinear complementarity equalities, enlarging the feasible set. As in the original, the single-stud capacity $T$ is applied after solving and is not a constraint of either program.

 Enlarging the feasible set can only lower the optimal value, so our program can only under-estimate how far from equilibrium the structure must be left. To validate the reformulation we cross-check our HiGHS solution against the commercial solver of StableLego~\citep{gurobi}. Table~\ref{tab:lp-agree} reports a three-way comparison. Running the commercial solver on the \emph{same} relaxation isolates whether our rewrite contains a modelling error; running it on the \emph{original} program isolates whether dropping the complementarity equalities changes any verdict. Neither produces a single disagreement, the certificate holds everywhere, optimal values agree to $6\times10^{-6}$, and solving all the states takes $8.8$\,s against $140.2$\,s.

All $70$ states are partial states of a single structure from the held-out test split, spanning $2$ to $55$ bricks: thirty lie along two complete disassembly trajectories of it, one from a greedy ordering and one from our planner, and forty are formed by deleting a random subset of its bricks. The random states are the point of the design, since trajectory states are almost all stable and a comparison made only on those would have nothing to discriminate. Because they come from one structure, this establishes agreement across a wide range of partial states and verdicts, not across structure categories.

\begin{table}[htbp]
\centering
\caption{Agreement between our linear program and the original, on $70$ partial states of one held-out test structure, $2$ to $55$ bricks. Both solvers run single-threaded with a fixed seed.}
\label{tab:lp-agree}
\small
\begin{tabular}{@{}ll@{}}
\toprule
Quantity & Result \\
\midrule
HiGHS against the commercial solver, same relaxation     & $70/70$ verdicts identical \\
HiGHS against the commercial solver, original program    & $70/70$ verdicts identical \\
complementarity certificate holds                        & $70/70$ \\
difference in optimal objective value                    & $\le 6\times10^{-6}$ \\
time to solve all $70$ states                            & $8.8$\,s against $140.2$\,s ($15.9\times$) \\
\bottomrule
\end{tabular}
\end{table}

\section{External Comparison}
\label{app:extbaselines}

This appendix details the baselines compared in \revise{Table~\ref{tab:baselines}} of the main paper: two world-model planners (H-WM and PhysWorld), one optimization (OR) solver, and two planner-based methods, ASAP~\citep{asap} and WorkBenchMark~\citep{workbenchmark}.

\subsection{H-WM (logical world model)}
H-WM~\citep{hwm} has no public code release, and its full setting assumes an RGB observation stream and a downstream vision-language-action policy, neither of which exists in our task, which is assembling a brick set on an integer stud grid with no vision. We therefore reimplement its \emph{logic-only} ablation (its Section~5.4), following the logical world-model design of its Section~4.1 exactly as described, and train it on our data. Training follows H-WM directly, which adopts the training formulation of Self-CriTeach~\citep{selfcriteach}: a full fine-tune of Qwen3-4B, learning rate $1\mathrm{e}{-5}$, global batch size $16$, cosine schedule, AdamW, trained for up to $10$k steps. \revise{Its results are reported in Table~\ref{tab:baselines} of the main paper.}

\subsection{PhysWorld (residual-PPO policy)}
PhysWorld~\citep{physworld} targets robot learning from generated video through a physical world model, a different task setting from ours. We port its only trainable component, a residual PPO policy (${\sim}367$k parameters), onto scenes built from our structures and train it on our data, while leaving out its video generation, monocular reconstruction, texturing, and pose estimation, which are specific to its own tasks. \revise{We evaluate it on the same $1{,}200$-structure set as the other world-model baselines; its success rate and paired comparison are reported in Table~\ref{tab:baselines} of the main paper.} It drives exact-oracle calls to our level ($0.54$ vs.\ our $0.52$ per brick) but does not learn physical feasibility: its mis-placement rate is $0.104$ per brick, $5.4\times$ ours ($0.019$).

\subsection{OR constraint solver}
\label{app:orsolver}
We implement an exact constraint solver for assembly-by-disassembly following the static blocking-graph formulation of Culbertson et al.~\citep{culbertson}, using a static blocking set with CP-SAT and a stability lazy no-good. The model chooses a partition of the bricks into rigidly-coupled groups of size at most five, together with an order over those groups. For a group $g$ placed at step $s[g]$, let $t[i]$ be the step of brick $i$. Let $x[g]$ be the indicator that selects group $g$, $y[g,o]$ the indicator that selects one minimal blocking set $B_o$ as the geometric condition, and $p[i,j]$ the ground-support parent pointer. The constraints are:
\begin{align}
&\text{AllDifferent}(s), \\
&\textstyle\sum_{g \ni i} x[g] = 1 && \forall i, \\
&x[g]=1 \Rightarrow t[i]=s[g] && \forall i \in g, \\
&\textstyle\sum_{o} y[g,o] = x[g], \\
&x[g]=1 \Rightarrow t[j] < s[g] && \forall j \in \big(\textstyle\bigcup_{i \in g}\text{above}(i)\big)\setminus g, \\
&y[g,o]=1 \Rightarrow t[j] < s[g] && \forall j \in B_o, \\
&\textstyle\sum_{j \in N(i)} p[i,j] + p[i,\perp] = 1, \\
&p[i,j]=1 \Rightarrow t[j] \ge t[i].
\end{align}
The objective is feasibility: any assignment satisfying all constraints within the budget. A complete solver must enumerate every connected subset of size at most five, and the number of such groups grows as $\Theta(n\,\bar d^{\,k-1})$, where $n$ is the number of bricks, $k\le5$ is the maximum group size, and $\bar d$ is the mean degree of the vertical contact graph. This up-front enumeration is the tradeoff our world-model rollout avoids by scoring candidates as they arise. On $120$ structures stratified by brick count, under the same $1800$\,s per-structure budget and the same exact ruler, counting timeouts as failures, the OR solver solves $40/120$ ($33.3\%$) against our $110/120$ ($91.7\%$); above $100$ bricks it solves only ${\sim}10\%$.

\subsection{Planner-based baselines: ASAP and WorkBenchMark}
\label{app:planner-baselines}
For ASAP~\citep{asap} we re-implement the method following its released code and retrain its GNN on the StableLego train split, using our exact LP stability labels. To separate the effect of distribution shift from the effect of the stability ruler, we report a ladder on the $75$ small structures ($9$--$32$ bricks) under our exact LP ruler: $2/75$ with the original released weights, $6/75$ when those weights are paired with our LP, $11/75$ when ASAP is retrained on our data, and $26/75$ when it is both retrained on our data and scored with our LP. For reference our planner solves $49/75$ at group\_max${=}1$ and $75/75$ at group\_max${\le}5$ on the same structures. Retraining reorders ASAP's search but still cannot clear the collision verification, which isolates the collision head and the look-ahead world model as the source of our advantage. For WorkBenchMark~\citep{workbenchmark} we re-implement its Algorithm~1 removal gate and preserve its full search procedure; both planner-based baselines operate in the single-brick action space they assume and are compared against Ours at group\_max${=}1$.

\section{Latent Dynamics vs.\ Re-Encoding the Next Structure}
\label{app:dyn-reencode}
The latent dynamics head takes exactly two inputs: the current anchor latent $z_i$ and the fixed featurization $\psi_i(a)$ of the removed brick relative to anchor $i$ (its occupancy footprint within the anchor's cube concatenated with its $11$-dimensional attribute vector). It does not read the next-state occupancy grid, exposed-face set, or per-brick contact counts. The alternative---re-encoding the true next structure for the affected anchors---must first materialize and maintain those global quantities. We compare the two directly: updating a latent through the dynamics head takes $0.34$\,ms, against $1.56$\,ms to produce the same update by re-encoding. The substitution is lossless: against a control that re-encodes the exact next structure, the two agree under a paired test with $p=1$, at $0.0175$ versus $0.0178$ misplacements per brick.

\end{document}